\documentclass{article}
\usepackage{arxiv}
\usepackage{inconsolata}
\usepackage{amsmath}

\usepackage[utf8]{inputenc} % allow utf-8 input
\usepackage[T1]{fontenc}    % use 8-bit T1 fonts
\usepackage{hyperref}       % hyperlinks
\usepackage{url}            % simple URL typesetting
\usepackage{fontawesome5}
\usepackage{booktabs}       % professional-quality tables
\usepackage{graphicx} 
\usepackage[table]{xcolor}
\usepackage{amsfonts}       % blackboard math symbols
\usepackage{nicefrac}       % compact symbols for 1/2, etc.
\usepackage{microtype}      % microtypography
\usepackage{xcolor}         % colors
\usepackage[most]{tcolorbox} 
\definecolor{tkcolor}{RGB}{235, 235, 255}
\usepackage{multicol}
\usepackage{multirow}
\usepackage{enumitem}
\usepackage{xspace}
\usepackage{subfigure}
\usepackage{listings}
\usepackage{xcolor}
\usepackage{enumitem}
\usepackage{parskip}
\usepackage{framed} % For framed environme
\usepackage{pifont}
\usepackage{makecell}
 \usepackage{titletoc}

\newcommand\FormalML{FormalML\xspace}

\newcommand{\cmark}{\ding{51}}%
\definecolor{xmark}{RGB}{175,175,175}
\newcommand{\xmark}{\textcolor{xmark}{\ding{55}}}%

\definecolor{keywordcolor}{rgb}{0.7, 0.1, 0.1}   % red
\definecolor{tacticcolor}{rgb}{0.0, 0.1, 0.6}    % blue
\definecolor{commentcolor}{rgb}{0.4, 0.4, 0.4}   % grey
\definecolor{symbolcolor}{rgb}{0.0, 0.1, 0.6}    % blue
\definecolor{sortcolor}{rgb}{0.1, 0.5, 0.1}      % green
\definecolor{attributecolor}{rgb}{0.7, 0.1, 0.1} % red

\lstdefinestyle{prompt}{
  language=lean_no_color,
  breakindent=0pt,
  basicstyle=\normalsize,
}
\lstdefinestyle{lean}{
  language=lean,
  frame=single,
  breakindent=0pt,
  basicstyle=\ttfamily
}
\lstdefinestyle{leaninline}{
  language=lean,
  frame=single,
  breakindent=0pt,
  basicstyle=\ttfamily,
}

\usepackage[utf8]{inputenc}
\lstdefinestyle{jsonstyle}{
    style=lean,  
    language=json,
    backgroundcolor=\color{lavender!20},
    mathescape=true, 
}

\definecolor{paleviolet}{HTML}{E1EEFC}
\definecolor{lightgrey}{RGB}{247, 247, 247}
\definecolor{darkgrey}{rgb}{0.5, 0.5, 0.5}
\definecolor{darkgreen}{rgb}{0, 0.5, 0}

\definecolor{delim}{RGB}{20,105,176}
\definecolor{numb}{RGB}{0,0,0}
\definecolor{string}{rgb}{0.6,0.08,0.08}
\definecolor{jsoncomment}{rgb}{0.5,0.5,0.5}

\lstdefinelanguage{json}{
    showspaces=false,
    showtabs=false,
    breaklines=true,
    breakatwhitespace=true,
    basicstyle=\ttfamily\small,
    upquote=true,
    morecomment=[l][\color{jsoncomment}]{\#},
    morestring=[b]",
    stringstyle=\color{string}\ttfamily,
    columns=[l]fullflexible,
    literate=
     *{0}{{{\color{numb}0}}}{1}
      {1}{{{\color{numb}1}}}{1}
      {2}{{{\color{numb}2}}}{1}
      {3}{{{\color{numb}3}}}{1}
      {4}{{{\color{numb}4}}}{1}
      {5}{{{\color{numb}5}}}{1}
      {6}{{{\color{numb}6}}}{1}
      {7}{{{\color{numb}7}}}{1}
      {8}{{{\color{numb}8}}}{1}
      {9}{{{\color{numb}9}}}{1}
      {\{}{{{\color{delim}{\{}}}}{1}
      {\}}{{{\color{delim}{\}}}}}{1}
      {[}{{{\color{delim}{[}}}}{1}
      {]}{{{\color{delim}{]}}}}{1},
}

\lstdefinestyle{jsonstyle}{
    language=json,
    backgroundcolor=\color{lavender!20}
}

\newenvironment{abstractenv}{
  \begin{tcolorbox}[
    colback=gray!10,    
    colframe=gray!50,   
    boxrule=0.5pt,       
    arc=2mm,             
    left=6pt, right=6pt,  
    top=6pt, bottom=6pt,
    coltitle=black,
    fonttitle={\bfseries}, title=Abstract
  ]
}{
  \end{tcolorbox}
}

\begin{document}

% Define and set date
\makeatletter
\def\icmldate#1{\gdef\@icmldate{#1}}
\icmldate{\today}
\makeatother

% Define page style
% \makeatletter
% \fancypagestyle{fancytitlepage}{
%   \fancyhead{}  % clear all headers
%   % \lhead{\includegraphics[height=1.5cm]{figures/logos/lamda.png}}
%   \rhead{\it \@icmldate}
%   \cfoot{}
% }
% \makeatother

% The \icmltitle you define below is probably too long as a header.
% Therefore, a short form for the running title is supplied here:

\icmltitle{FormalML: A Benchmark for Evaluating Formal Subgoal Completion in Machine Learning Theory}
% \maketitle
% \begin{icmlauthorlist}
% \mbox{Xiao-Wen Yang$^{\,1\,*\,\dagger\,}$},
% \mbox{Zihao Zhang$^{\,1\,*}$}, 
% \mbox{Jianuo Cao$^{\,1\,}$}, 
% \mbox{Zhi Zhou$^{\,1\,}$}, 
% \mbox{Zenan Li$^{\,2\,}$}, 
% \mbox{Lan-Zhe Guo$^{\,1\,}$}, 
% \mbox{Yuan Yao$^{\,1\,}$}, 
% \mbox{Taolue Chen},
% \mbox{Yu-Feng Li},
% and \mbox{Xiaoxing Ma$^{\,1\,}$\textsuperscript{\faEnvelope}}
% \end{icmlauthorlist}
\begin{icmlauthorlist}
\mbox{Xiao-Wen Yang$^{1,2*}$\textsuperscript{\href{mailto:yangxw@lamda.nju.edu.cn}{\faEnvelope}}},
\mbox{Zihao Zhang$^{1*}$\textsuperscript{\href{mailto:zihaozhang@smail.nju.edu.cn}{\faEnvelope}}}, 
\mbox{Jianuo Cao$^{1,3}$}, 
\mbox{Zhi Zhou$^1$}, 
\mbox{Zenan Li$^{4}$}, 
\mbox{Lan-Zhe Guo$^{1,3}$}, 
\mbox{Yuan Yao$^{1}$}, 
\mbox{Taolue Chen$^{5}$},
\mbox{Yu-Feng Li$^{1,2\dagger}$\textsuperscript{\href{mailto:liyf@lamda.nju.edu.cn}{\faEnvelope}}},
and \mbox{Xiaoxing Ma$^{1\dagger}$\textsuperscript{\href{mailto:xxm@nju.edu.cn}{\faEnvelope}}}
\end{icmlauthorlist}

$^{1}$State Key Laboratory of Novel Software Technology, Nanjing University \\
$^{2}$School of Artificial Intelligence, Nanjing University \\
$^{3}$School of Intelligence Science and Technology,
Nanjing University\\
$^{4}$Department of Computer Science, ETH Zurich, Switzerland \\
$^{5}$School of Computing and Mathematical Sciences, Birkbeck, University of London, UK

$^{*}$ Equal Contribution \quad $^{\dagger}$ Corresponding Author

% \icmlcorrespondingauthor{\{le-y22, zq-chen23\}@mails.tsinghua.edu.cn, gaohuang@tsinghua.edu.cn}

% \printAffiliations{}
\vskip .3in

% \printNotice{} % otherwise use the standard text.

\begin{abstractenv}
    
Large language models (LLMs) have recently demonstrated remarkable progress in formal theorem proving.
Yet their ability to serve as practical assistants for mathematicians—filling in missing steps within complex proofs—remains underexplored.
We identify this challenge as the task of \textit{subgoal completion}, where an LLM must discharge short but nontrivial proof obligations left unresolved in a human-provided sketch.
To study this problem, we introduce \textbf{FormalML}, a Lean 4 benchmark built from foundational theories of machine learning. Using a translation tactic that converts procedural proofs into declarative form, we extract 4,937 problems spanning optimization and probability inequalities, with varying levels of difficulty. FormalML is the first subgoal completion benchmark to combine premise retrieval and complex research-level contexts. Evaluation of state-of-the-art provers highlights persistent limitations in accuracy and efficiency, underscoring the need for more capable LLM-based theorem provers for effective subgoal completion.

        \makebox[\linewidth]{
        ~~
    }
   \makebox[\linewidth]{
\includegraphics[height=0.8em]{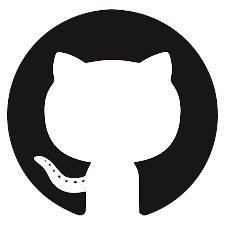} \textit{Github}: \href{https://github.com/njuyxw/FormalML}{ \ttfamily https://github.com/njuyxw/FormalML}
}
\makebox[\linewidth]{
\includegraphics[height=0.8em]{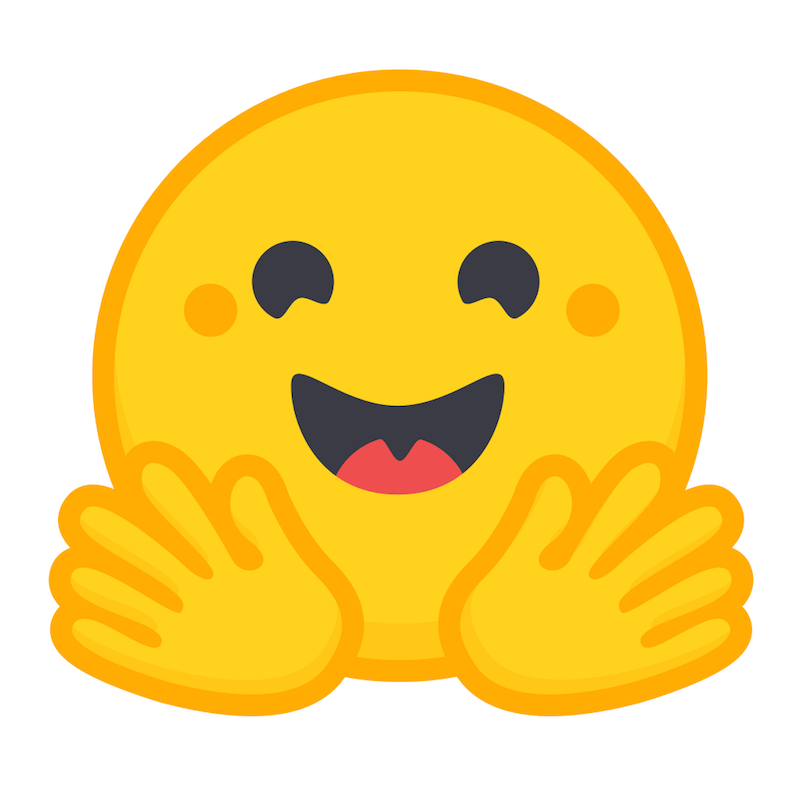} \textit{HuggingFace}: \href{https://huggingface.co/datasets/zzhisthebest/FormalML}{ \ttfamily https://huggingface.co/datasets/zzhisthebest/FormalML}
}
\end{abstractenv}

\section{Introduction}

% 1. success in a competition-level problem
% 2. fail in research-level, and why
% 3. first step: subgoal completion.
% 4. benchmark, but lacks 
% 5. key observation

Recent advancements in large language models (LLMs) have showcased their impressive performance in formal theorem proving~\citep{ren2025deepseek, kiminap, bfsprover},
particularly culminating in silver-medal level performance at the 2025 International Mathematical Olympiad~\citep{chen2025seed}. 
Nevertheless, their effectiveness as copilots for mathematicians in tackling research-level problems~\citep{yang2024formal} remains far from satisfactory.
For example, the recent Equational Theories Project~\citep{EquationalTheoriesProject2024} reported that ``LLMs did not provide useful suggestions beyond what the human participants could already propose''  in most of the difficult cases.

\begin{figure}[t]
    \centering
    \includegraphics[width=0.99\textwidth]{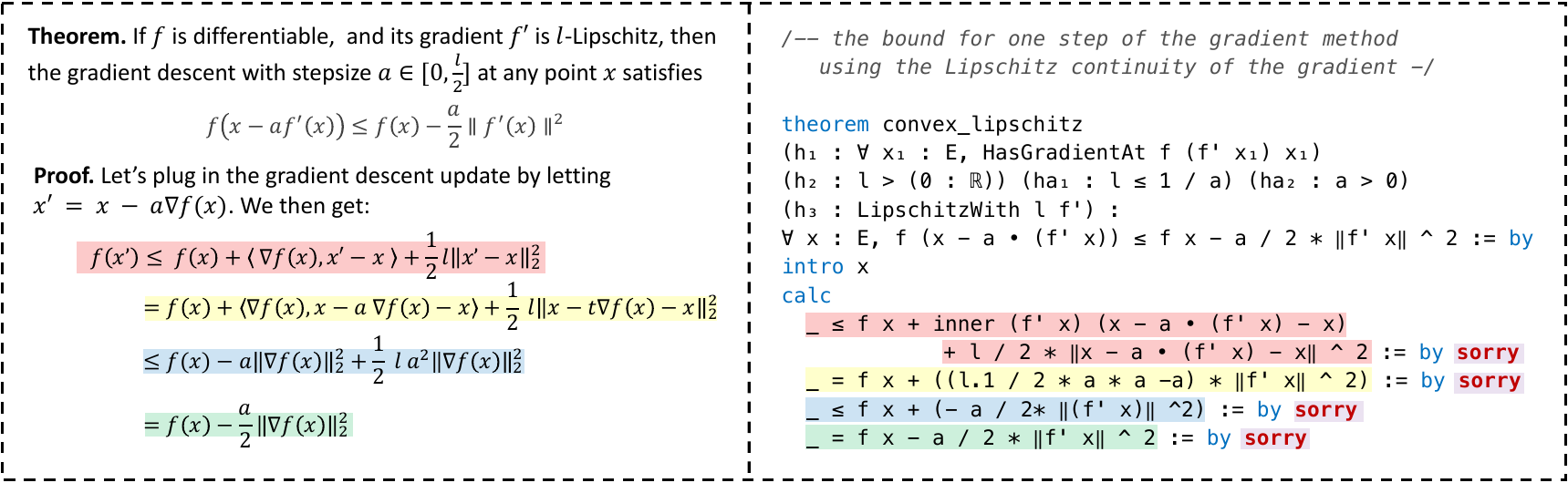}
    % \vspace{-2mm}
    \caption{The informal and formal versions of the statement and proof regarding the sufficient decrease in gradient descent. Both the formal statement and proof can be accurately formalized from the natural language version. However, to verify each step in Lean 4, there are unproven holes (notated as \lstinline[style=leaninline]{sorry}) that need to be further completed by existing automated theorem provers.}
    \label{fig:motivation}
    % \vspace{-5mm}
\end{figure}

In this paper, rather than attempting to generate a complete proof for such research-level problems, we investigate an intermediate milestone, which we refer to as \emph{subgoal completion}.
To elaborate, human experts specify the problem in natural language, formalize the theorem, and outline the high-level proof structure, while AI systems assist by filling in the missing technical details~\citep{jiang2022draft}.
Figure~\ref{fig:motivation} illustrates this process using the proof of gradient descent convergence~\citep{li2024formalization, li2025formalizationsmooth};
The expert provides the informal statement and the main reasoning steps.
At the natural language level, the proof appears complete; however, once formalized, it remains incomplete, as indicated by the \lstinline[style=leaninline]{sorry} placeholders. 
This gap arises because the theorem prover cannot automatically bridge each reasoning step.
Subgoal completion precisely aims to bridge this gap, aligning the informal and formal proofs.

Despite its natural role as an interface between informal and formal reasoning, subgoal completion still lacks dedicated benchmarks. 
Existing evaluations are built around competition-style problems and emphasize full-proof generation, whereas our target lies in research-level theorems and the task of completing intermediate proofsteps. 
Moreover, while proof-generation benchmarks primarily measure the LLM performance by proving success rate, subgoal completion poses a different challenge: handling complex proof contexts efficiently without resorting to verbose or speculative reasoning chains. 
In practice, a useful prover must therefore strike a balance between accuracy and conciseness. 
To achieve these goals, we make the following contributions:

\begin{itemize}[leftmargin=*,nosep]
\setlength\itemsep{0.42em}
\item \textbf{A Lean 4 translation tactic for subgoal extraction.} 
We propose a symbolic strategy for the dataset construction.
Specifically, we implemented a new Lean 4 tactic that extracts subgoal problems from procedural-style proofs. 
By adapting the length of the proof segments during extraction, we further utilize this tactic to generate new problems with varying levels of proving difficulty. 

\item \textbf{The \FormalML benchmark.} We narrow the scope to the foundational theory of machine learning (ML) and establish a new benchmark in this domain.
Our motivation is twofold.
First, AI agents are becoming increasingly central to automating scientific discovery~\citep{romera2024mathematical}, particularly within machine learning research~\citep{lu2024ai, yamada2025ai, gottweis2025towards}.
Hence, automated theorem proving plays a key role in guaranteeing the soundness of derived results.
Second, automation tools that verify theoretical correctness can substantially ease the burden on human reviewers~\citep{xuneurips, pineau2021improving}, especially given the rapidly growing volume of ML conference submissions.
A well-designed benchmark would accelerate the development of such tools. 
To this end, building on two Lean 4 libraries—Optlib~\citep{li2024formalization} and FoML~\citep{sonoda2025lean}, we establish a dataset of 4,937 subgoal completion problems.
\item \textbf{Systematic evaluation of LLM-based provers on \FormalML}.
We evaluate state-of-the-art LLM-based provers on \FormalML, highlighting fundamental limitations in accuracy, token-efficiency, and premise retrieval. Our results show that while models such as DeepSeek-Prover-V2~\citep{ren2025deepseek} improve retrieval capabilities, overall performance drops sharply on higher-difficulty problems. Moreover, chain-of-thought prompting, though effective in natural language reasoning, fails to improve proof completion and often reduces efficiency in this context. Additionally, we find expert iteration to be a effective training approach on \FormalML. These insights highlight the need for further development and refinement in LLM-based theorem provers to better support mathematicians in their work. 
\end{itemize}

% To identify and highlight this gap, we focus on the theorem proving of machine learning theory and propose a new benchmark \FormalML for practical proof completion.

\section{Related Work}
% update 更新最新文章：Seed-Prover: Deep and Broad Reasoning for Automated Theorem Proving + seed另外一篇
\textbf{LLM-based Theorem Proving.}
In recent years, the rapid development of LLMs has significantly advanced research on formal theorem proving~\citep{survey}. Current mainstream approaches can be broadly divided into two categories. The first category employs tree-search strategies, including best-first search~\citep{leandojo,bfsprover,internlm,gptf}, Monte Carlo tree search~\citep{deepseekprover1_5,dt-solver,mcts,carts}, and others~\citep{lean-star}. 
Although these methods align with the Markov Decision Process (MDP) framework of theorem proving, their step-by-step tactic generation and frequent interaction with the Lean environment result in low search efficiency. 
The second category is whole-proof generation~\citep{stp,goedel,leanabell,kiminap,ren2025deepseek,chen2025seed}, where LLMs directly generate complete proofs, usually through multiple rollouts verified in Lean to identify the final proof. 
This approach has demonstrated notable advantages. For instance, DeepSeek-Prover-V2~\citep{ren2025deepseek} and Kimina-Preview~\citep{kiminap} leverage natural language-aided Long-CoT reasoning to produce formal proofs, achieving breakthroughs in competition-level tasks. 
However, the performance of existing LLM-based theorem provers in supporting practical formal proving has not yet been fully evaluated.

\textbf{Formal Theorem Proving Benchmarks.}
The current mainstream Lean formal theorem proving benchmarks can be categorized into three primary types based on problem characteristics. The first type focuses on Olympiad-level high school mathematics competition problems, exemplified by: the MiniF2F~\citep{minif2f} dataset, which comprises 244 challenging problems from competitions such as AMC, AIME, and IMO; and the PutnamBench benchmark~\citep{putnambench}, featuring problems derived from the Putnam Mathematical Competition. The second type targets undergraduate-level mathematical tasks, including ProofNet~\citep{proofnet} (containing analysis and algebra problems at the undergraduate level) and MinCTX~\citep{hu2024minictx} (theorems sourced from real-world Lean projects and textbooks). Additionally, hybrid benchmarks like FormalMATH~\citep{yu2025formalmath} cover theorem proving across domains ranging from high school to undergraduate mathematics. Another category, exemplified by Leandojo~\citep{leandojo}, directly constructs datasets from the Mathlib library.
Notably, existing benchmarks either operate within simple contexts or do not require premise retrieval, and all involve full proof generation tasks. In contrast, our proposed FormalML benchmark specializes in subgoal completion tasks, a design better aligned with practical applications where LLM-based theorem provers assist humans in practical theorem proving. A comprehensive comparative analysis of benchmarks is presented in Table \ref{tab:benchmark}.

\begin{table}[h]
\centering
% \vspace{-4mm}
\caption{Comparison of existing Lean 4 benchmarks.}
\label{tab:benchmark}
\resizebox{1.0\textwidth}{!}{
\begin{tabular}{lccccc}
\toprule
\textbf{Benchmark} & \textbf{\# Problems} & \textbf{Type}  & \textbf{Premise} & \textbf{\makecell{Complex\\Context}} & \textbf{\makecell{Subgoal\\Completion}}\\ 
\midrule
MiniF2F~\citep{minif2f} & $244$ & Olympiad & \xmark & \xmark & \xmark \\
PutnamBench~\citep{putnambench} & $522$ & Olympiad & \xmark & \xmark& \xmark  \\
ProofNet~\citep{proofnet} & $186$ & Undergraduate (UG) & \xmark & \cmark & \xmark \\
FormalMATH~\citep{yu2025formalmath} & $5,560$ & Olympiad \& UG  & \xmark & \cmark & \xmark \\
LeanDojo~\citep{leandojo} & $2000$ & Mathlib & \cmark & \xmark& \xmark \\
MiniCTX~\citep{hu2024minictx} & $762$ & UG  &\cmark & \cmark& \xmark  \\
\rowcolor{gray!20} FormalML (Ours) & $4,937$ &  UG & \cmark & \cmark & \cmark \\
\bottomrule
\end{tabular}}
\end{table}
\section{\FormalML}

\subsection{Subgoal Extraction}
\begin{figure}[t]
  \centering
  \includegraphics[width=0.99\linewidth]{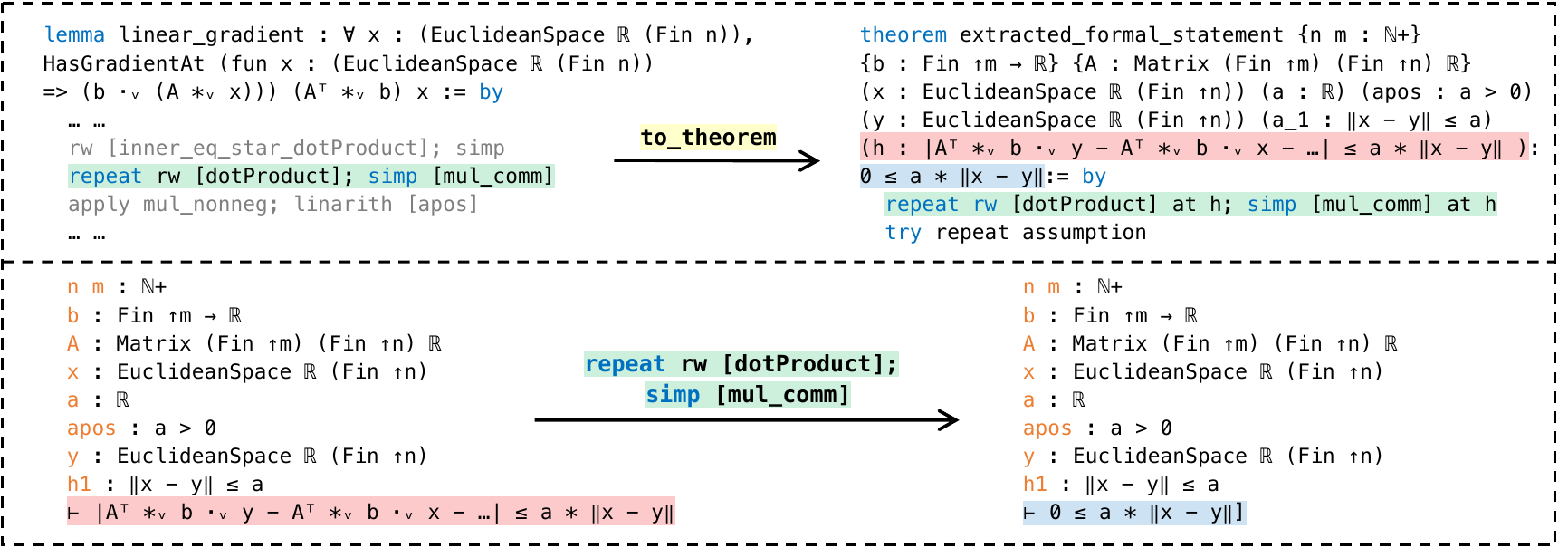}
  % \vspace{-2mm}
  \caption{An example of the \lstinline[style=leaninline]{to_theorem} tactic illustrates its functionality. When applied to the tactic \lstinline[style=leaninline]{repeat rw [dotProduct]; simp [mul_comm]}, it captures pre- and post-execution proof states, abstracts their transition, and then synthesizes a subgoal.}
\label{fig:to_theorem}
% \vspace{-2mm}
\end{figure}

With the growing availability of formalized libraries in machine learning theory, we aim to leverage these resources for dataset construction.
A key challenge, however, lies in extracting subgoals from procedural-style proof scripts~\citep{winograd1975frame, wiedijk2012synthesis}.
Unlike declarative-style proofs~\citep{syme1997declare}, which progress by iteratively introducing and proving intermediate subgoals, procedural-style proofs instead transform the proof state step by step.
As a result, existing goal-extraction tactics such as Lean 4’s \lstinline[style=leaninline]{extract_goal} can only capture the overall goal of a proof state, but not the finer-grained intermediate subgoals corresponding to each reasoning step.
This limitation is particularly pressing as most proofs in current libraries are written in the procedural style. 
For example, in the proof of the lemma \lstinline[style=leaninline]{linear_gradient} 
(see Figure~\ref{fig:to_theorem} and Appendix~\ref{appendix:linear_gradient_proof}), procedural scripts such as \lstinline[style=leaninline]{rw} and \lstinline[style=leaninline]{simp} span nine lines, yet declarative statements (e.g., \lstinline[style=leaninline]{have}) are entirely absent.

Fortunately, in human-written Lean 4 proofs, we observe a consistent pattern: \emph{although proofs are expressed procedurally as sequences of tactics, each proof line typically corresponds to a single reasoning step identified by the human author.}
In practice, human experts tend to first complete the reasoning informally and subsequently encode it in tactics.
Accordingly, a single line frequently chains multiple tactics until the intended reasoning step has been fully realized, at which point a new line is introduced.
Consider the running example in Figure~\ref{fig:to_theorem}, the expert combines two tactics (i.e., \lstinline[style=leaninline]{repeat rw [dotProduct]} and \lstinline[style=leaninline]{simp [mul_comm]}) within a single line to ensure that the resulting proof state aligns with the underlying informal reasoning.
This observation motivates us to analyze human-written proofs at the line level.

We implement a customized Lean 4 tactic \lstinline[style=leaninline]{to_theorem}, which automatically encodes line-level, procedural proofsteps into a new subgoal, with the corresponding proofsteps serving as their proofs.
As shown in Figure~\ref{fig:to_theorem}, the tactic operates by recording the proof states immediately before and after executing the procedural steps \lstinline[style=leaninline]{repeat rw [dotProduct]; simp [mul_comm]}.
It then inserts the prior state as hypotheses and the subsequent state as the goal, thereby deriving a new theorem.
This theorem effectively isolates a subgoal of the original proof, which can then be discharged directly by replaying the original procedural tactics.

The tactic \lstinline[style=leaninline]{to_theorem} extracts the subgoal from a single line.
We can naturally extend it to proof-segment extraction, where a sequence of tactics spanning multiple lines is first converted into the declarative format, and then its subgoal is extracted.
For the running example \lstinline[style=leaninline]{linear_gradient} in Figure~\ref{fig:to_theorem}, which totally comprises a nine-line proof, we can, in principle, extract up to nine single-line subgoals; four non-overlapping two-line subgoals; and so forth.

\subsection{Data Curation}

% \subsection{Data Sources}

We categorize machine learning theories into two main types: optimization theory and probability theory. 
The former typically analyzes the convergence of optimization algorithms (e.g., gradient descent), while the latter addresses error bounds of machine learning models (e.g., Hoeffding's inequality).
To develop \FormalML, we expand and extract data based on two projects: \textbf{Optlib} \citep{li2024formalization} and \textbf{FoML} \citep{sonoda2025lean}. 
For each project, we designate the top-level theorems (e.g., the convergence of gradient descent) for subgoal extraction, 
while preserving lower-level lemmas (e.g., Lipschitz continuity) as a local library for premise retrieval.
We utilize the \lstinline[style=leaninline]{to_theorem} tactic to extract the subgoals, and ultimately compile 4,937 theorems into \FormalML.

Optlib~\citep{li2024formalization, li2025formalization, li2025formalizationsmooth} is a project started in September 2023 that formalizes a wide range of optimization algorithms, including gradient descent (GD), subgradient method (SubGD), proximal gradient descent (PGD), Nesterov acceleration method (NAG), block coordinate descent (BCD), and alternating direction minimization method (ADMM).
Building on Optlib, we extend and refine theorems related to these algorithms, and update some proofs to ensure compatibility.

FoML~\citep{sonoda2025lean} is a recent project initiated in March 2025, primarily focused on formalizing the generalization error bound using Rademacher complexity. It also includes some important probability inequalities, such as the expectation inequality, the bounded differences inequality, and McDiarmid's inequality. Furthermore, we formalize Hoeffding’s lemma, the Bennett inequality, and the Bernstein inequality into the library to enhance its comprehensiveness.

Each theorem in \FormalML is stored in JSON format (See Appendix ~\ref{appendix:json format}), including:
% \begin{enumerate}[leftmargin=2em, label=(\arabic*)]
\begin{enumerate}[leftmargin=2em, label=(\arabic*), nosep, topsep=0pt, itemsep=0pt]
\item Source location metadata (file path and line number coordinates for theorem extraction).
\item The formal Lean 4 theorem statement.
\item Required module imports and namespace declarations.
\item The complete tactic sequence constituting the original proof.
\item Retrieved proof-relevant premises from the mathematical context.
\end{enumerate}
% For each theorem, the proof context is complete, allowing for independent testing of each extracted theorem, thus facilitating subsequent parallel evaluations. 
% We release both the dataset and the evaluation code as open source.
This encapsulation ensures a complete proof context for each theorem, enabling independent evaluation.  We release both the dataset and the evaluation code as open source.
\subsection{Benchmark Challenges}
At first glance, subgoal completion may appear simpler than the proof generation for complex competition problems, in which LLM-based theorem provers have demonstrated their excellent performance.
However, we identify that the proofstep generation task remains particularly challenging for LLM-based theorem provers, attributable to the following three key reasons.
\begin{itemize}[leftmargin=*,nosep]
\setlength\itemsep{0.42em}
\item \textbf{Complex proof context.} 
  In practice, users often rely on automated theorem provers to close subgoals within lengthy and complex proofs. 
  This requires LLMs to comprehend all accumulated hypotheses and intricate goals to generate accurate proofs. However, current provers pay little attention to complex contextual reasoning.

  \item \textbf{Premise retrieval.} 
Accurately retrieving relevant premises from both local and global libraries is crucial, as many proofs depend on interdependent lemmas~\citep{hu2024minictx}. Thus, effective retrieval is a prerequisite for practical subgoal completion. However, current LLM-based provers still perform inadequately in this regard \cite{leandojo}.
  
  \item \textbf{Overthinking and efficiency.} 
Unlike competition problems that demand long chains of reasoning, subgoal completion is often repetitive and involves relatively straightforward reasoning.
LLMs trained for complex competitions may \emph{overthink} such tasks, exploring unnecessary inference paths that hinder performance~\citep{chiang2024over,sui2025stop}.
Moreover, efficiency is crucial: it is impractical to expend substantial computational resources on proving subgoals that require only brief, direct proofs.
Balancing reasoning depth with computational cost is thus a key challenge for practical subgoal completion.
\end{itemize}

\subsection{Benchmark Statistics}

\begin{table}[t]
\centering
% \vspace{-1.5em}
\caption{Statistics of theorems in \FormalML across various machine learning theories.}
\resizebox{1.0\textwidth}{!}{
\begin{tabular}{lrrrrrrr|c}
\toprule
\textbf{Optimization} & {GD} & {SubGD} & {PGD} & {NAG} & {BCD} & {ADMM} & {Other} & Total \\
{\# Problems} & 211 & 331 & 388 & 528 & 433 & 1,016 & 0 & 2,907 \\
\midrule
\textbf{Probability} & {Exp} & {Bennett} & {McDiarmid} & {Rademacher} & {Hoeffding} & {Measure} & {Other} & Total\\
{\# Problems} & 100 & 136 & 642 & 615 & 52 & 315 & 170 & 2,030 \\
\bottomrule
\end{tabular}}
\label{tab:data_dist}
\vspace{-1em}
\end{table}

\begin{figure*}[t]
\centering
\includegraphics[width=0.99\textwidth]{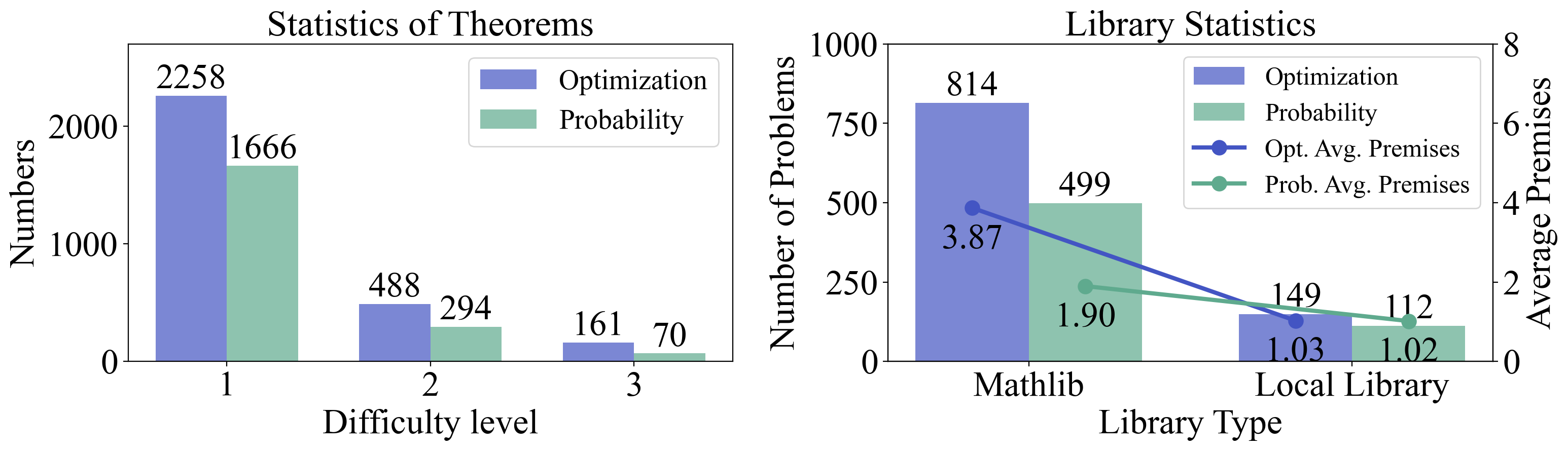}
\vspace{-2mm}
\caption{Statistics of \FormalML theorems: (Left) distribution by proving difficulty; (Middle) retrieval sources and averages; (Right) average number of retrieved premises by difficulty.}
\label{fig:stat}
\vspace{-5mm}
\end{figure*}
\textbf{Summary.} 
We establish \FormalML by systematically expanding and extracting theorems from two Lean 4 libraries: {Optlib} and {FoML}. The resulting benchmark comprises 4,937 unique theorems, each derived from a single proof step or proof segment within the top-level theorems presented in these libraries.
For the extracted theorems, we categorize them according to their corresponding top-level theorems, as illustrated in Table~\ref{tab:data_dist}. The results indicate that the benchmark encompasses a diverse range of theories in machine learning. The number of extracted theorems associated with each top-level theorem depends on the length of the corresponding proof. On average, hundreds of theorems can be successfully extracted from each top-level theorem.

\textbf{Proving difficulty.} 
To enable a fine-grained evaluation of LLM-based theorem provers, we categorize theorems based on their proving difficulty.
Specifically, 
since the proofs in \FormalML are mostly procedural style, we can use the proof length as a metric of proving difficulty~\citep{zhang2024selene}. 
We define three levels of proving difficulty, corresponding to proof lengths of 1, 3, and 5, respectively, and report the statistics of theorems across these levels in the left panel of Figure~\ref{fig:stat}. In total, there are 3,924 theorems at difficulty level 1 and 1,013 theorems at higher difficulty levels.

\textbf{Retrieval difficulty.}  
We further examine the retrieval difficulty within \FormalML. In total, 2,049 theorems require explicit premise retrieval. These theorems are divided into two categories based on the source of the retrieved premises: Mathlib and the local library. The middle panel of Figure~\ref{fig:stat} reports the number of theorems in each category as well as the average number of premises. Furthermore, the right panel of Figure~\ref{fig:stat} summarizes the average number of retrieved premises required at different difficulty levels, showing that more complex proofs generally demand greater retrieval effort.

\newtcolorbox{findings}[1][]{
	width=\columnwidth,
	colback = tkcolor, 
	colframe = tkcolor, 
boxsep=0pt,left=10pt,right=10pt,top=5pt,bottom=5pt,
	fontupper=\linespread{0.9}\selectfont,
	title=#1}

% \newpage
\section{Experiments}
% \vspace{-3mm}
\begin{table}[!t]
\caption{Performance comparison of LLM-based theorem provers on \FormalML. The best results are in bold, and the second-best are underlined.}
\centering
 \resizebox{1.0\textwidth}{!}{
\begin{tabular}{l c c ccc}
\toprule
   \multirow{2.5}{*}{\textbf{Method}}  & \multirow{2.5}{*}{\textbf{\makecell{Model\\Size}}} & \multirow{2.5}{*}{\textbf{\makecell{Sample\\Budget}}} & \multicolumn{3}{c}{\textbf{Pass@$K$ (\%)}} \\
        \cmidrule(lr){4-6}
        & & & \textbf{Optim.} & \textbf{Prob.} & \textbf{All} \\
\toprule
\multicolumn{6}{c}{\textit{Best-First Tree Search Methods}} \\[0.5mm]
\multirow{2}{*}{Reprover~\citep{leandojo}} & \multirow{2}{*}{229M} &$4\times20$&  $$21.80$$ & $$16.15$$ & $$19.48$$ \\
& &$8\times50$&  $$27.82$$ & $$20.44$$ & $$24.79$$ \\
\midrule
\multirow{2}{*}{BFS-Prover~\citep{bfsprover}} & \multirow{2}{*}{7B} &$4\times20$&  $$24.45$$ & $$19.31$$ & $$22.26$$ \\
& &$8\times50$&  $$27.62$$ & $$23.84$$ & $$25.31$$ \\
\toprule
\multicolumn{6}{c}{\textit{Whole-Proof Generation Methods}} \\[0.5mm]
\multirow{4}{*}{Kimina-Prover-Preview-7B~\citep{kiminap}} & \multirow{4}{*}{7B}
    &1  & $$17.99$$ & $$11.77$$ & $$15.43$$ \\
  &  &4  & $$30.75$$ & $$23.69$$ & $$27.85$$ \\
  &   &16 & $$42.31$$ & $$38.13$$ & $$40.59$$ \\
  &   &32 & $$47.09$$ & $$43.79$$ & $$45.74$$ \\
\midrule
\multirow{4}{*}{Goedel-Prover~\citep{goedel}} & \multirow{4}{*}{7B}
    &1  & $$8.22$$ & $$13.50$$ & $$10.39$$ \\
  &   &4  & $$21.32$$ & $$28.97$$ & $$24.47$$ \\
  &   &16 & $$37.29$$ & $$41.28$$ & $$38.93$$ \\
  &   &32 & $$44.34$$ & $$46.11$$ & $$45.07$$ \\
\midrule
\multirow{4}{*}{Goedel-Prover-V2-8B~\citep{lin2025goedelv2}} & \multirow{4}{*}{7B}
    &1  & 20.85 & 19.31 &20.21  \\
  &   &4  & 34.61 &33.15  &34.01  \\
  &   &16 & 45.13 & 43.35 & 44.40 \\
  &   &32 & 49.67 & 48.67&49.26  \\
\midrule
\multirow{4}{*}{Leanabell-Prover~\citep{leanabell}} & \multirow{4}{*}{7B}
    &1  & $$24.11$$ & $$24.09$$ & $$24.10$$ \\
  &   &4  & $$47.33$$ & $$43.30$$ & $$45.68$$ \\
  &   &16 & $$58.00$$ & $$50.84$$ & $$55.05$$ \\
&   &32 & $$61.23$$ & $$53.99$$ & $$58.07$$ \\
\midrule
\multirow{4}{*}{STP~\citep{stp}} & \multirow{4}{*}{7B}
    &1  & $$28.45$$ & $$24.83$$ & $$26.96$$ \\
   &  &4  & $$51.01$$ & $$46.06$$ & $$48.98$$ \\
   &  &16 & $$62.13$$ & \underline{$$57.73$$} & $$60.32$$ \\
   &  &32 & $$\textbf{65.19}$$ & $$\textbf{60.39}$$ & $$\textbf{63.21}$$ \\
\midrule
\multirow{4}{*}{DeepSeek-Prover-V1.5~\citep{deepseekprover1_5}} & \multirow{4}{*}{7B}
    &1  & $$24.29$$ & $$17.29$$ & $$21.41$$ \\
   &  &4  & $$43.31$$ & $$37.39$$ & $$40.88$$ \\
   &  &16 & $$57.48$$ & $$52.56$$ & $$55.46$$ \\
   &  &32 & $$61.51$$ & $$56.00$$ & $$59.25$$ \\
\midrule
\multirow{4}{*}{DeepSeek-Prover-V2 (noCoT)~\citep{ren2025deepseek}} & \multirow{4}{*}{7B}
    &1  & $$17.16$$ & $$16.40$$ & $$16.85$$ \\
   &  &4  & $$37.98$$ & $$37.09$$ & $$37.61$$ \\
   &  &16 & $$59.79$$ & $$52.76$$ & $$56.90$$ \\
   &  &32 & \underline{$$65.08$$} & \underline{$$57.73$$} & \underline{$$62.06$$} \\
\midrule
\multirow{4}{*}{DeepSeek-Prover-V2 (CoT)~\citep{ren2025deepseek}} & \multirow{4}{*}{7B}
    &1  & $$16.23$$ & $$21.58$$ & $$18.43$$ \\
   &  &4  & $$25.73$$ & $$32.56$$ & $$28.54$$ \\
   &  &16 & $$33.54$$ & $$39.36$$ & $$35.93$$ \\
   &  &32 & $$37.39$$ & $$42.51$$ & $$39.50$$ \\
\bottomrule
\end{tabular}}
\label{tab:main-results}
% \vspace{-2mm}
\end{table}
In this section, we conduct a comprehensive evaluation of state-of-the-art LLM-based theorem provers using our \FormalML benchmark. Through systematic experiments, we identify several limitations and phenomena characterizing current theorem provers in practical subgoal completion tasks.
\subsection{Evaluating LLM-based Theorem Provers on \FormalML}
\paragraph{Theorem Provers} 
We primarily focus on the two most effective types of LLM-based provers:
\begin{itemize}[leftmargin=*,nosep]
\setlength\itemsep{0.42em}
    \item \textbf{Best-First Tree-Search (BFS) Methods:} Each node in the search tree corresponds to a proof state, and a heuristic scoring function assigns priorities to nodes. The BFS algorithm is employed to explore the search space and derive the final proof. We evaluate two models in this framework: Reprover~\citep{leandojo} and BFS-Prover~\citep{bfsprover}.
    \item \textbf{Whole-Proof Generation Methods:} These types of models generate a complete formal proof in a single pass directly from the problem description, eliminating the need for additional state transitions or search procedures. We consider the following SOTA models: STP~\citep{stp}, Goedel-Prover~\citep{goedel}, Godel-Prover-V2~\citep{lin2025goedelv2}, Leanabell-Prover~\citep{leanabell},  Kimina-Prover-Preview-7B~\citep{kiminap}, DeepSeek-Prover-V1.5~\citep{deepseekprover1_5} and DeepSeek-Prover-V2~\citep{ren2025deepseek}.
\end{itemize}

\paragraph{Metrics \& Evaluation} We employ the Pass@$K$ metric to evaluate theorem provers, where $K$ represents the computational budget. This metric quantifies the proportion of problems for which at least one valid proof is discovered within the top $K$ generated attempts. For BFS methods, $K=E\times T$, where $E$ represents the number of tactics generated per expansion, and $T$ denotes the number of expansion iterations. Unlike prior studies that predominantly focus on competition-level theorem proving—characterized by highly challenging problems and lengthy proofs, thereby requiring large values of $K$, this work concentrates on the research level subgoal completion. Consequently, our benchmark emphasizes low computational budgets, corresponding to small values of $K$ ($\leq 8\times50$ for BFS methods and $\leq 32$ for whole-proof generation methods).

\paragraph{Experimental Details} For BFS methods, we implement and conduct interactive search based on the LeanDojo~\citep{leandojo} framework. For whole-proof generation methods, we employed vLLM~\citep{vllm} to generate proofs and evaluated their correctness using kimina-lean-server~\citep{kiminaleanserver}, which provides high-throughput Lean 4 verification. All experiments strictly adhered to the published parameter configurations and prompts for each model. All computations were performed on NVIDIA H800 GPUs.

\paragraph{Main Results} In Table \ref{tab:main-results}, we show the performances of current theorem provers on our \FormalML benchmark.  Our experiments demonstrate that current whole-proof generation models have preliminarily acquired the capability to solve \FormalML problems when increasing sample budgets, STP attaining the highest Pass@32 score of 63.21\%. However, their Pass@$1$ performances remains at a relatively low level (peaking at merely 26.96\%), falling short of the practical requirements for assisting mathematicians in completing Lean formal proofs. Notably, the BFS tree search strategy, despite consuming more computational resources, does not yield significant performance improvements (remaining below 30\%). 
Additionally, we observe that: 1) the performance ranking of models exhibits inconsistencies with the results from the miniF2F benchmark, suggesting that certain models exhibit varying degrees of overfitting to competition-level elementary theorem-proving tasks;
2) recent long-CoT provers (e.g. Kimina-Preview and Deepseek-Prover-V2) incorporating natural language assistance underperform on our 
\FormalML benchmark (with pass@32 consistently below 50\%), presenting a stark contrast to their strong performance on miniF2F.

\begin{figure}[t]
\centering
% \subfigure{
% \hspace{-0.5cm}
\subfigure{
\includegraphics[width=0.48\textwidth]{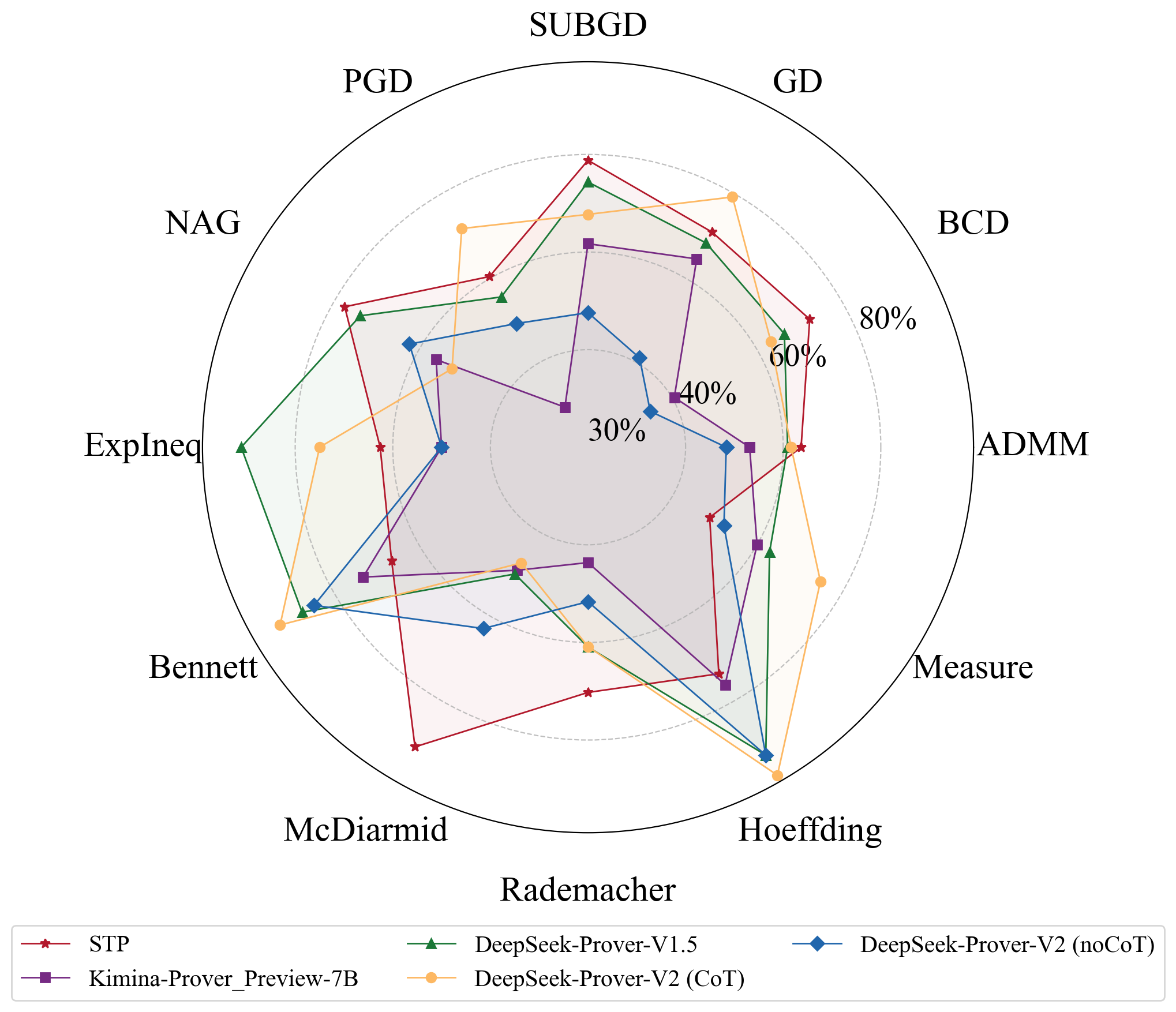}}
\subfigure{
\includegraphics[width=0.48\textwidth]{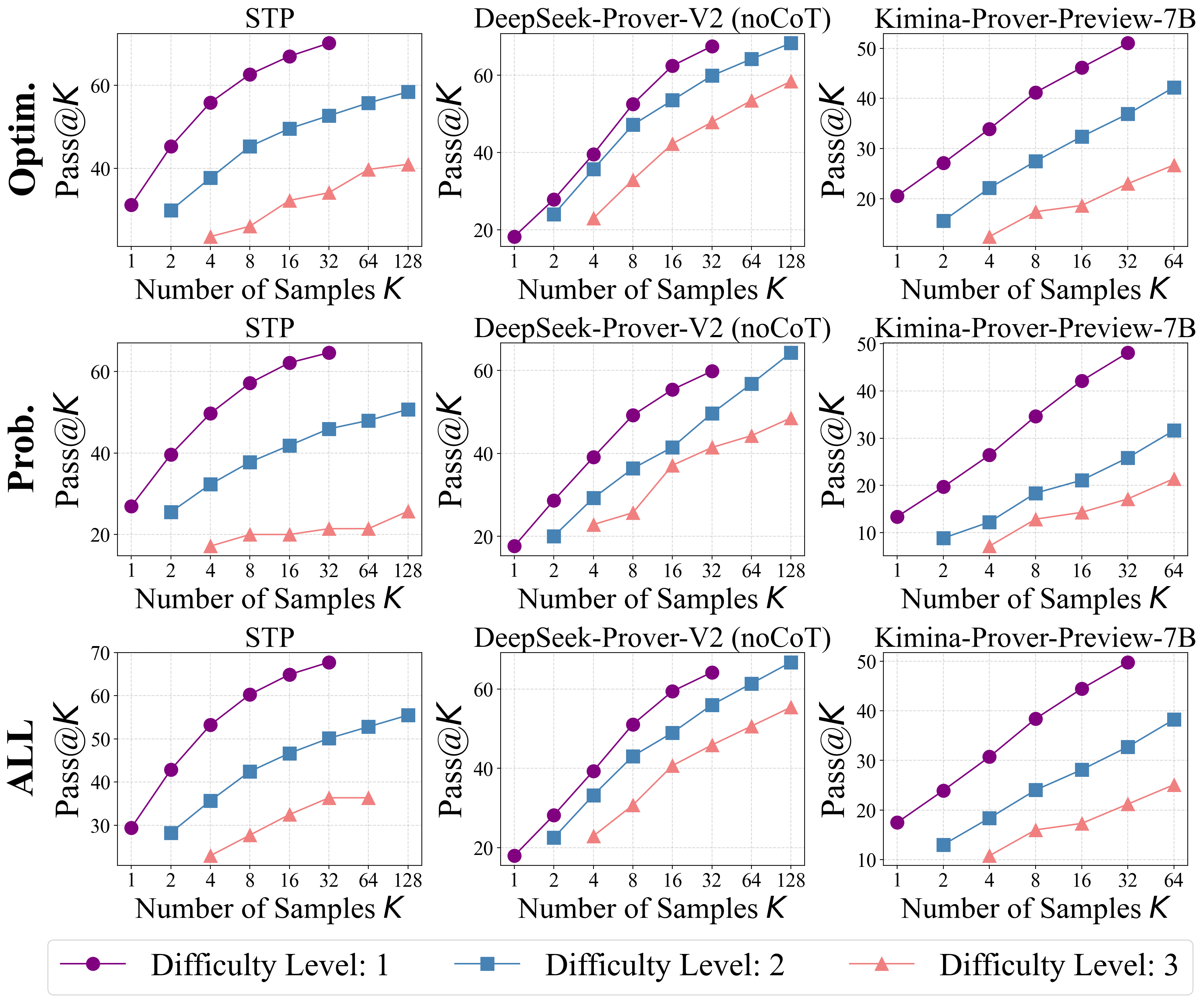}}
\centering
% \vspace{-2mm}
\caption{The left figure presents results of pass rate across various specific problem domains, while the right figure shows the performances under different difficulty levels.}
\label{fig:ld}
% \vspace{-5mm}
\end{figure}

\paragraph{Performance Distribution} We present the distributional results of performance across various specific problem domains (e.g., the proof of Hoeffding's inequality). 
The experimental results presented in left panel of Figure \ref{fig:ld} demonstrate that the performance of existing models varies across different specialized areas. For instance, in the probability split, STP achieves significantly superior results on Mcdiarmid, far surpassing other methods. On Hoeffding, Deepseek-Prover-V2 (noCoT) delivers the best performance.

\begin{findings}
\textbf{Finding 1:} Existing LLM-based theorem provers are inadequate as practical tools for assisting mathematicians in achieving subgoal completion \textit{under low computational budgets}. Moreover, the performance of LLM-based theorem provers exhibits significant disparities across different specialized areas within \FormalML.
\end{findings}

% \begin{findings}
% \textbf{Finding 2:} 
% \end{findings}

\subsection{Evaluating Retrieval Effectiveness on \FormalML}
While writing practical proof, humans often need to retrieve premises from local theorem libraries or Mathlib.
The experiments are conducted on the retrieval subset of our benchmark, with the following specific design: for each proposition to be proved, we randomly sample candidate premises from a combined premise library comprising Mathlib and local premises, mixing in actually correct theorems to ultimately form a retrieval set containing $M$ candidate theorems. All candidate theorems are presented in a structured format within the model's context (the complete prompt template is provided in Appendix \ref{app:prompts}). Considering the current context window limitations of LLM-based provers, this study adopts two experimental configurations ($M=10$ and $M=20$) for comparative analysis. Our study intentionally omits a comparative evaluation of embedding-based retrieval models.  This design choice stems from the consideration that using the same model for both retrieval and proof generation to enhance efficiency in the subgoal completion task.

Results are demonstrated in Table \ref{tab:re}.
We observe that most models exhibit improved performance when provided with candidate premises, yet their effectiveness declines as $M$ increases (indicating higher difficulty).    Although STP demonstrates strong performance across the full benchmark, it underperforms in premise retrieval.    We attribute this to the absence of such data in its training set. We identify a several representative case in Appendix \ref{case:re}.
In contrast, DeepSeek-Prover-V2 achieves robust results under both CoT and noCoT settings, with a high performance gain of approximately 10\%  at $M=10$.    This indicates its strong retrieval capability, which we think stems from its training on extensive natural language reasoning data, thereby enhancing its general reasoning ability.

\begin{findings}
\textbf{Finding 2:} Existing LLM-based provers often struggle with premise retrieval in complex proofs, resulting in suboptimal performance. However, models like DeepSeek-Prover-V2 demonstrate superior retrieval capabilities, attributable to their strong general reasoning abilities.
\end{findings}

\subsection{Evaluating \FormalML of Different Difficulty Level}
We evaluate the performance of existing whole-proof generation models at varying difficulty levels on \FormalML. A larger $K$ is used to assess problems of higher difficulty. As shown in right of Figure \ref{fig:ld}, experimental results indicate that model performance, both for the optimization and the probability split, decreases as difficulty increases. Notably, STP still achieves the highest results under difficulty levels 3 and 5, with Pass@128 scores of 55.5\% and 33.36\%, respectively. Results of other models are shown in Appendix \ref{app:fig-diff}.

\begin{table}[t]
\centering
\caption{Performances of retrieval and then proof. Relative improvements compared to $M=0$ are shown in {green} (increase) or {red} (decrease).}
\resizebox{1.0\textwidth}{!}{
\begin{tabular}{l *{6}{c}}
\toprule
\multirow{2}{*}{Method} & \multicolumn{2}{c}{$\mathbf{M=0}$}& \multicolumn{2}{c}{$\mathbf{M=10}$} & \multicolumn{2}{c}{$\mathbf{M=20}$} \\
\cmidrule(lr){2-3} \cmidrule(lr){4-5} \cmidrule(lr){6-7}
 & Pass@16(\%) & Pass@32(\%) & Pass@16(\%) & Pass@32(\%) & Pass@16(\%) & Pass@32(\%) \\
\midrule
STP &54.56 &57.72 &52.23 \scriptsize{\textcolor{red}{(-2.33)}} &56.43 \scriptsize{\textcolor{red}{(-1.29)}} &52.17 \scriptsize{\textcolor{red}{(-2.39)}} &56.04 \scriptsize{\textcolor{red}{(-1.68)}} \\
Goedel-Prover &32.97&39.04&34.58 \scriptsize{\textcolor{green}{(+1.61)}}&41.05 \scriptsize{\textcolor{green}{(+2.01)}}&33.35 \scriptsize{\textcolor{green}{(+0.38)}}&39.50 \scriptsize{\textcolor{green}{(+0.46)}}\\
Goedel-Prover-V2-8B &45.24&51.26&45.05 \scriptsize{\textcolor{red}{(-0.19)}}&49.90 \scriptsize{\textcolor{red}{(-1.36)}} &46.09 \scriptsize{\textcolor{green}{(+0.85)}}&50.36 \scriptsize{\textcolor{red}{(-0.90)}}\\
Leanabell-Prover &46.67&50.42&47.45 \scriptsize{\textcolor{green}{(+0.78)}}&50.87 \scriptsize{\textcolor{green}{(+0.45)}}&46.93 \scriptsize{\textcolor{green}{(+0.26)}}&50.29 \scriptsize{\textcolor{red}{(-0.13)}}\\
Kimina-Prover-Preview-7B &34.71&39.56&34.65 \scriptsize{\textcolor{red}{(-0.06)}}&39.69 \scriptsize{\textcolor{green}{(+0.13)}}&35.49 \scriptsize{\textcolor{green}{(+0.78)}}&39.75 \scriptsize{\textcolor{green}{(+0.19)}}\\
Deepseek-Prover-V1.5&49.45 &52.94 &46.86 \scriptsize{\textcolor{red}{(-2.59)}} &53.59 \scriptsize{\textcolor{green}{(+0.65)}}  &42.79 \scriptsize{\textcolor{red}{(-6.66)}} &49.45 \scriptsize{\textcolor{red}{(-3.49)}}\\
Deepseek-Prover-V2 (noCoT) &53.46 &58.37 &58.82 \scriptsize{\textcolor{green}{(+5.36)}} &69.29 \scriptsize{\textcolor{green}{(+10.92)}} &57.40 \scriptsize{\textcolor{green}{(+3.94)}} &65.80 \scriptsize{\textcolor{green}{(+7.43)}}\\
Deepseek-Prover-V2 (CoT) &34.39&37.94&44.34 \scriptsize{\textcolor{green}{(+9.95)}}&47.71 \scriptsize{\textcolor{green}{(+9.77)}}&42.73 \scriptsize{\textcolor{green}{(+8.34)}}&46.74 \scriptsize{\textcolor{green}{(+8.80)}}\\
\bottomrule
\end{tabular}}
\label{tab:re}
% \vspace{-5mm}
\end{table}

\begin{findings}
\textbf{Finding 3:} Current LLM-based theorem provers tend to exhibit significantly degraded performance when handling higher-difficulty problems in \FormalML.
\end{findings}
\subsection{Evaluating Theorem Prover Efficiency on \FormalML}

In formal theorem proving, efficient provers are crucial for practical applications. Recent studies demonstrate that RL based long-CoT reasoning has been successfully integrated into systems such as DeepSeek-Prover-V2, Kimina-Prover and Goedel-Prover-V2. These models employ deep natural-language reasoning before generating formal proofs. While they exhibit strong performance on competition-level mathematical problems, our experiments reveal no significant advantage on the \FormalML benchmark (See Table \ref{tab:main-results}). Moreover, these models generally suffer from low reasoning efficiency.  To address the trade-off between reasoning efficiency and accuracy in \FormalML scenarios, we use a novel evaluation metric—Efficiency-Weighted Accuracy (EWA@$K$)—defined as:  
$$
\text{EWA@}K = \text{Pass@}K \times \frac{100}{\text{Response Length}}
$$
This metric balances performance and output efficiency by normalizing the proof success rate against the response length. 
To the right of Figure \ref{fig:gra}, we compare the pass rates and efficiency of eight whole-proof generation models on \FormalML. The results show that the three long-COT models generate the highest average number of response tokens, far exceeding other models. However, their performance is inferior to others, resulting in the lowest EWA@32 score. In contrast, STP achieves both the shortest output length and the highest pass rate on the \FormalML benchmark, leading to the highest EWA@32 score.  

\begin{findings} \textbf{Finding 4:} 
% While CoT improves proof generation accuracy, it significantly reduces practical efficiency for the subgoal completion task.
Long-CoT demonstrates no significant improvement in the pass rates of the subgoal completion task while simultaneously incurring substantial computational overhead.
\end{findings}

\subsection{Evaluating \FormalML with Expert Iteration}

We further explored the performance of expert iteration on \FormalML.
We extracted 92,815 problems from five repositories (mathlib, PrimeNumberTheoremAnd, PFR, PhysLean, and scilean) using the \lstinline[style=leaninline]{to_theorem} tactic, of which 88,174 were used for expert iteration training. For each problem, we generated 8 candidate proofs and sampled the correct ones for training. We performed 1 round of expert iteration on DeepSeek-Prover-V2 (noCoT) and Goedel-Prover-V2-8B. Our Training used the AdamW optimizer with a learning rate of 1e-5, cosine learning rate schedule, warmup ratio of 0.1, batch size of 8 with gradient accumulation steps of 2 (effective batch size 16), and 3 epochs, with bf16 precision and DeepSpeed ZeRO for distributed optimization. All experiments were conducted using GPU acceleration, and evaluation was carried out on the \FormalML dataset.

\begin{figure}[t]
\centering
% \subfigure{
% \hspace{-0.5cm}
\subfigure{
\includegraphics[width=0.48\textwidth]{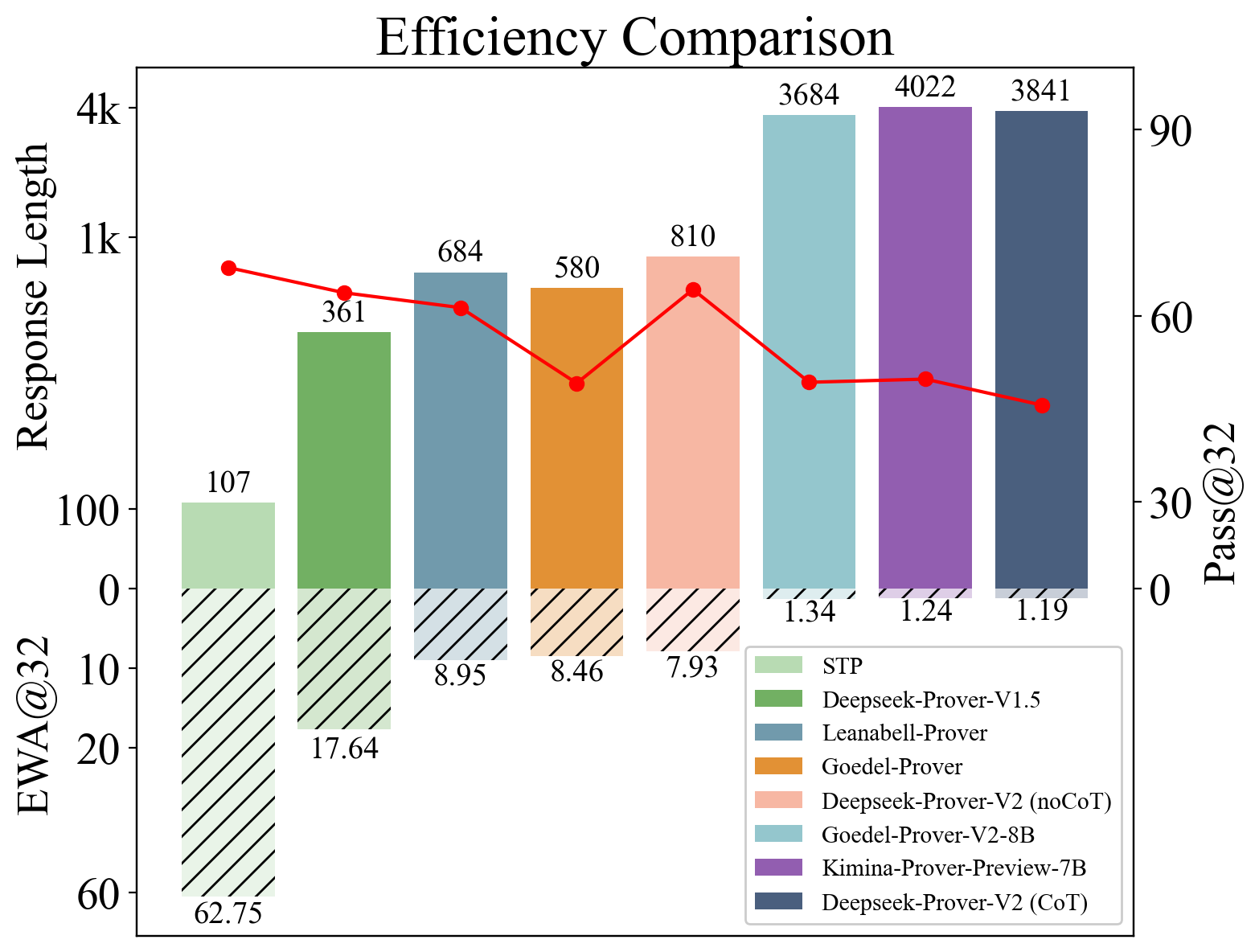}}
\subfigure{
\includegraphics[width=0.48\textwidth]{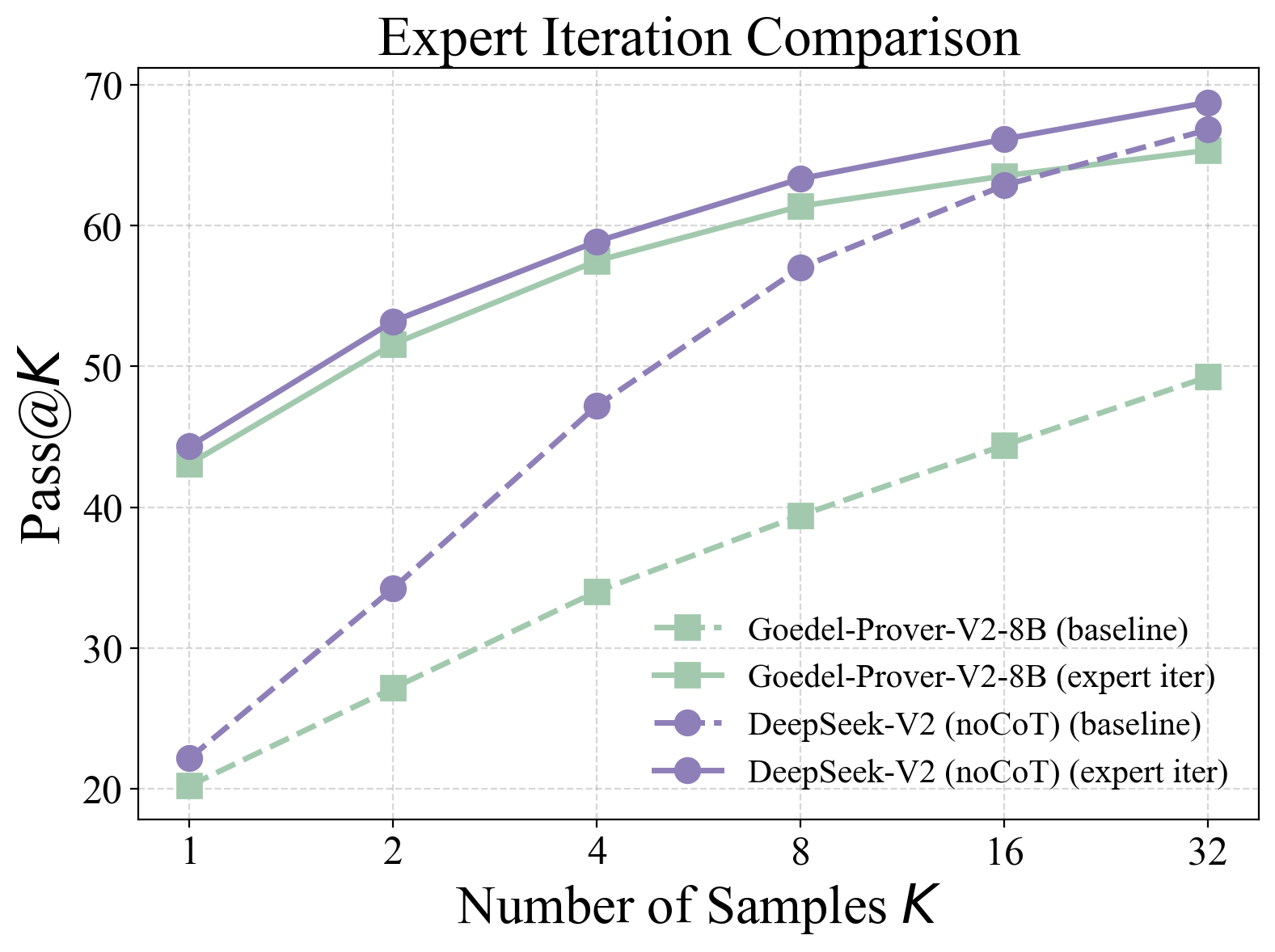}}
\centering
% \vspace{-2mm}
\caption{The left figure illustrates the efficiency comparison among current whole-proof generation provers, while the right shows performances before and after expert iteration.}
\label{fig:gra}
% \vspace{-5mm}
\end{figure}

Results are shown in the right panel of Figure \ref{fig:gra}. The experimental results indicate a substantial improvement in performance after expert iteration, especially in Pass@1 ((~20\%)). This suggests that expert iteration holds potential for enhancing subgoal completion tasks.

\begin{findings} \textbf{Finding 5:} 
It is demonstrated that expert iteration enhances performance on \FormalML and shows potential for improving subgoal completion capabilities.
% \vspace{-1mm}
\end{findings}
%

% Training used the AdamW optimizer with a learning rate of 1e-5, cosine learning rate schedule, warmup ratio of 0.1, batch size of 8 with gradient accumulation steps of 2 (effective batch size 16), and 3 epochs, with bf16 precision and DeepSpeed ZeRO for distributed optimization. All experiments were conducted using GPU acceleration, and evaluation was carried out on the FormalML dataset.

\section{Conclusion}
% We present FormalML, a benchmark designed to evaluate LLMs as collaborative assistants in formal proof completion for machine learning theory. We develop a proof-style translation tool that converts procedural proofs into a declarative format, improving clarity and facilitating structured problem extraction. By systematically curating and expanding existing theorem libraries, we construct a dataset of 4,767 formal problems. Our evaluation demonstrates that current LLM-based theorem provers remain limited in premise retrieving and generating proof steps within complex theorem contexts. We hope FormalML will drive progress toward more robust and efficient LLM-based theorem provers, ultimately advancing the automation of formal research in machine learning theory.
In this paper, we identify subgoal completion as a critical yet understudied task for helping mathematicians working on research-level problems.    We introduce \FormalML, the first benchmark specifically designed for this task in formal theorem proving.    Using a Lean 4 translation tactic, we systematically extract 4,937 subgoal problems from foundational machine learning theory, covering optimization and probability.    Evaluations of state-of-the-art LLM-based provers demonstrate persistent limitations in accuracy, efficiency, and retrieval within complex proof contexts.
Moreover, expert iteration has proven to be a promising direction for advancing subgoal completion.
We anticipate that \FormalML will catalyze the development of more capable provers to support subgoal completion.

\paragraph{Limitations and future work} Our current benchmark primarily focuses on a subset of main domains related to machine learning theory.  In the future, we aim to expand into more relevant fields.  Additionally, we plan to formalize certain aspects of ML theory using Lean, with the goal of encouraging more researchers to explore the use of LLM-based methods for assisting in formal theorem proving.

\newpage
\bibliography{main}
\bibliographystyle{icml2025}

%%%%%%%%%%%%%%%%%%%%%%%%%%%%%%%%%%%%%%%%%%%%%%%%%%%%%%%%%%%%

\newpage
\appendix

\newpage
\appendix
\startcontents[parts] 
\printcontents[parts]{l}{1}{\section*{Appendix}} 
\newpage
\section{Data Format Specification}
\label{appendix:json format}
\label{app:data_format}
The \FormalML{} dataset stores each theorem in a structured JSON format. Below is a detailed schema with field descriptions.
 \begin{lstlisting}[style=jsonstyle, xleftmargin=10pt, backgroundcolor=\color{white}]
 {
   "filename": string,
   "line": int,
   "tactic_state_before": string,
   "tactic": string,
   "tactic_state_after": string,
   "goal": string,
   "theorem_header": string,
   "formal_statement": string,
   "full_formal_statement": string,
   "retrieval": [
     {
       "library": string,
       "definition": string
     }
   ]
 }
 \end{lstlisting}

 Based on the file \texttt{FoML/FoML/ForMathlib/Probability/Moments.lean} (Line 98), we further provide an illustrative example for some typical entries. 
In this line, the corresponding proof tactic is \lstinline[style=leaninline]{apply aemeasurable_expt_hX}. 
By extracting the pre- and post-tactic states, we can formulate the corresponding subgoal as follows:
\begin{tcolorbox}[breakable, enhanced jigsaw, top=-5pt, bottom=-5pt]
\begin{lstlisting}[style=lean, frame=none, mathescape=true]
theorem extracted_formal_statement_27.{u_1}
{$\Omega$ : Type u_1}
{$m$ : MeasurableSpace $\Omega$}
{$\mu$ : Measure $\Omega$}
[inst : IsFiniteMeasure $\mu$]
($a$ $b$ : $\mathbb{R}$)
{$X$ : $\Omega$ → $\mathbb{R}$}
(hX : AEMeasurable $X$ $\mu$)
(h : $\forall^m$ ($\omega$ : $\Omega$) $\partial\mu$, $X$ $\omega$ ∈ Set.Icc $a$ $b$) :
let $e$ := fun $t$ $\omega$ => $\exp(t \cdot X \omega)$;
$\forall$ ($t'$ : $\mathbb{R}$), AEStronglyMeasurable ($e$ $t'$) $\mu$ := sorry
\end{lstlisting}
\end{tcolorbox}
Regarding the retrieval entry, it records the library (e.g., \texttt{FoML}), and specifies the definition of the retrieved lemma. For example:
\begin{tcolorbox}[breakable, enhanced jigsaw, top=-5pt, bottom=-5pt]
\begin{lstlisting}[style=lean, frame=none, mathescape=true]
lemma aemeasurable_expt
{$X$ : $\Omega$ → $\mathbb{R}$}
($t$ : $\mathbb{R}$)
(hX : AEMeasurable $X$ $\mu$) :
AEStronglyMeasurable (fun $\omega$ ↦ $\exp(t \cdot X(\omega))$) $\mu$
\end{lstlisting}
\end{tcolorbox}

\newpage
\section{The Full Proof of Lemma \texttt{linear\_gradient}}
\label{appendix:linear_gradient_proof}
\begin{tcolorbox}[breakable, enhanced jigsaw, top=-5pt, bottom=-5pt]
\begin{lstlisting}[style=lean, frame=none, mathescape=true]
private lemma linear_gradient : 
∀ x : (EuclideanSpace $\mathbb{R}$ (Fin n)),
HasGradientAt (fun x : (EuclideanSpace $\mathbb{R}$ (Fin n)) => 
(b $\bullet_v$ (A $\ast_{v}$ x)))(A$^T$ $\ast_{v}$ b) x := by
  intro x
  rw [HasGradient_iff_Convergence_Point]
  intro a  apos
  use a ; use apos
  intro y _
  rw [dot_mul_eq_transpose_mul_dot, 
    dot_mul_eq_transpose_mul_dot, ← dotProduct_sub]
  rw [EuclideanSpace.inner_eq_star_dotProduct]; simp
  repeat rw [dotProduct]; simp [mul_comm]
  apply mul_nonneg; linarith [apos]; apply norm_nonneg
\end{lstlisting}
\end{tcolorbox}

\section{Additional Results}
\subsection{Error Analysis}
This section presents a detailed breakdown of error types from additional experiments conducted on three different theorem provers: DeepSeek-Prover-V1.5, DeepSeek-Prover-V2 (CoT), and Goedel-Prover. The results are categorized into three main types: \texttt{has\_error} (Lean code with execution errors), \texttt{is\_valid\_with\_sorry} (code that verifies but contains the sorry tactic), and \texttt{is\_valid\_no\_sorry} (correctly verified proofs). These proportions were calculated for both pass@16 and pass@32 metrics across two different problem domains: Optimization and Probability.
The tables below provide a comprehensive overview of these results, allowing for a direct comparison of the performance characteristics of each prover.
\begin{table*}[t]
\centering
\caption{Error type analysis for different theorem provers under pass@16 and pass@32.}
\label{tab:error_analysis}
\resizebox{\textwidth}{!}{
\begin{tabular}{llccc}
\toprule
\textbf{Setting} & \textbf{Model} & \textbf{has\_error} & \textbf{is\_valid\_with\_sorry} & \textbf{is\_valid\_no\_sorry} \\
\midrule
\multirow{3}{*}{Pass@16: Optimization} 
 & DeepSeek-Prover-V1.5   & 74.80 & 0.02 & 25.16 \\
 & DeepSeek-Prover-V2 (CoT) & 63.16 & 10.68 & 15.49 \\
 & Goedel-Prover          & 90.08 & 0.00 &  9.92 \\
\midrule
\multirow{3}{*}{Pass@16: Probability} 
 & DeepSeek-Prover-V1.5   & 80.04 & 0.04 & 19.89 \\
 & DeepSeek-V2 (CoT) & 61.94 & 7.65 & 22.77 \\
 & Goedel-Prover          & 85.20 & 0.00 & 14.80 \\
\midrule
\multirow{3}{*}{Pass@32: Optimization} 
 & DeepSeek-Prover-V1.5   & 75.09 & 0.02 & 24.87 \\
 & DeepSeek-Prover-V2 (CoT) & 62.98 & 10.76 & 15.49 \\
 & Goedel-Prover          & 90.11 & 0.00 &  9.89 \\
\midrule
\multirow{3}{*}{Pass@32: Probability} 
 & DeepSeek-Prover-V1.5   & 79.88 & 0.04 & 20.04 \\
 & DeepSeek-V2 (CoT)& 61.82 & 7.61 & 22.96 \\
 & Goedel-Prover          & 85.23 & 0.00 & 14.77 \\
\bottomrule
\end{tabular}}
\end{table*}

The results indicate that long-CoT models such as DeepSeek-V2-CoT tend to output "sorry" more frequently, but demonstrate lower rates of Lean errors. Conversely, Goedel-Prover seldom outputs "sorry" tactic but exhibit higher Lean error frequencies. 
\subsection{Results Under Varying Difficulty Levels}
\label{app:fig-diff}
\begin{figure*}[h]
\centering
\includegraphics[width=0.9\textwidth]{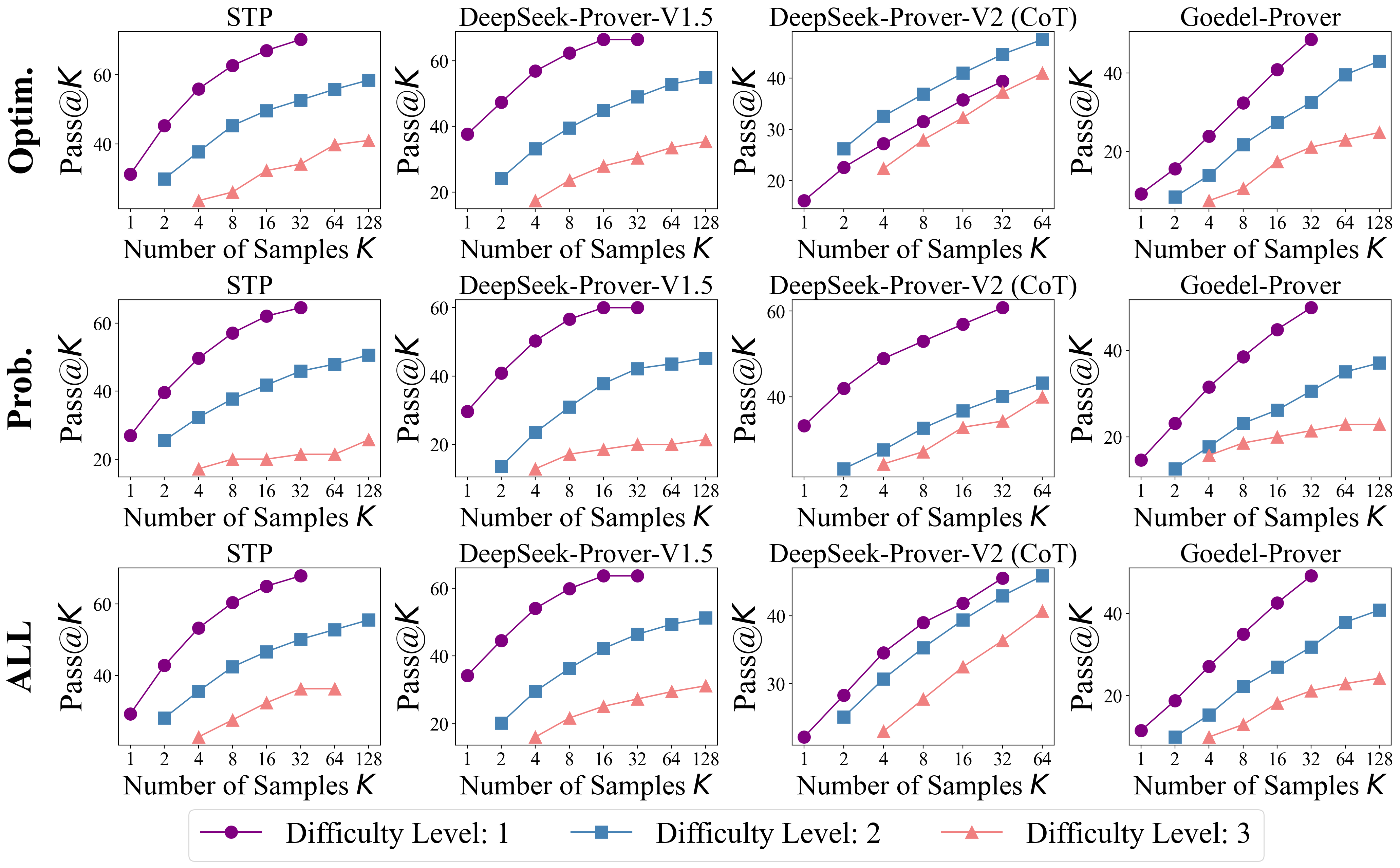}
\caption{Results of pass rate under varying difficulty levels (Part 1).}
\label{fig:all_diff1}
\end{figure*}

\begin{figure*}[h]
\centering
\includegraphics[width=0.9\textwidth]{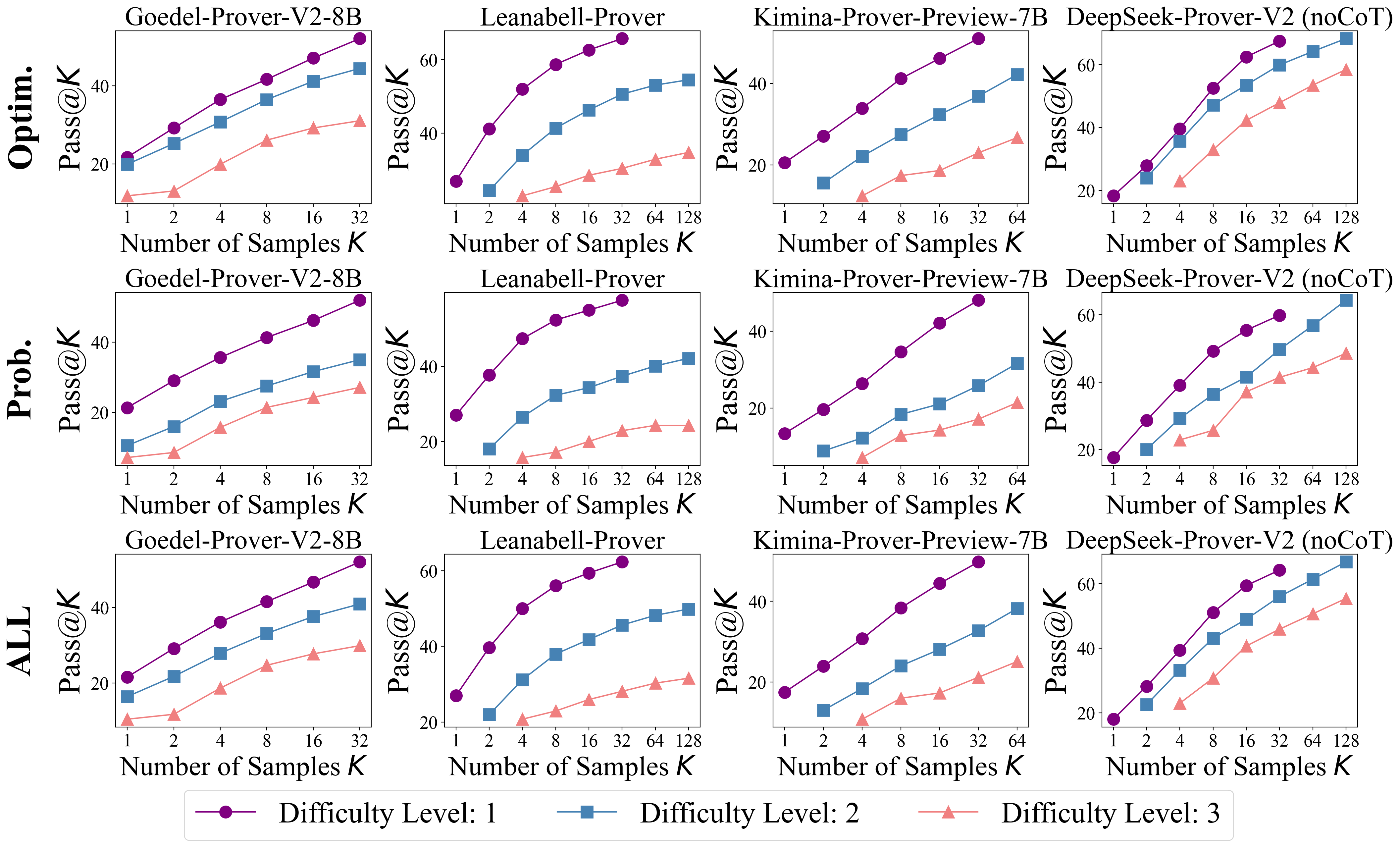}
\caption{Results of pass rate under varying difficulty levels (Part 2).}
\label{fig:all_diff2}
\end{figure*}

We supplement the experimental results for all eight whole-generation models across different difficulty levels, as presented in Figure \ref{fig:all_diff1} and Figure \ref{fig:all_diff2}. The results demonstrate a statistically significant trend where all models exhibit performance degradation as the difficulty level increases. Among all evaluated models, DeepSeek-Prover-V2 demonstrates the most modest performance decline.
% \begin{figure*}[h]
% \centering
% \includegraphics[width=\textwidth]{tex/Figs/granularity_plot_grid_all.png}
% \caption{Results of pass rate under varying difficulty levels.}
% \label{fig:all_diff}
% \end{figure*}

\section{Prompts for Retrieval}
\label{app:prompts}
For premise retrieval, we designed the following prompt, using DeepSeek-Prover-V2 as an illustrative example.
% For premise retrieval, we designed the following prompt, using DeepSeek-Prover-V2 as an illustrative example.
% retreival prompts
% An example prompt for retrieving of DeepSeekV2 (CoT)

\begin{tcolorbox}[top=-5pt, bottom=-5pt, title=The prompt for premise retrieval of DeepSeek-Prover-V2 (CoT)]
\begin{lstlisting}[style=prompt, basicstyle=\small]
You can use some of the following lemmas or theorems: {all_definitions}
Complete the following Lean 4 code: 
{formal_statement}
Before producing the Lean 4 code to formally prove the given theorem, provide a detailed proof plan outlining the main proof steps and strategies.
The plan should highlight key ideas, intermediate lemmas, and proof structures that will guide the construction of the final formal proof.
\end{lstlisting}
\end{tcolorbox}

\section{Case Study}

\subsection{Cases of Retrieval}
\label{case:re}
We present a case of STP where the problem was initially solved without retrieval, but errors emerged after candidate premises were provided.

\begin{tcolorbox}[breakable, enhanced jigsaw, top=-5pt, bottom=-5pt,title=Problem]
\begin{lstlisting}[style=lean, frame=none, mathescape=true, basicstyle=\small]
theorem extracted_formal_statement_162 
{n m : ℕ+} 
{A : Matrix (Fin ↑m) (Fin ↑n) ℝ}
{μ : ℝ} 
{μpos : 0 < μ} 
{Ane0 : A ≠ 0}  
(x : EuclideanSpace ℝ (Fin ↑n)) 
(a : x ∈ Set.univ) 
(ε : ℝ) 
(εpos : ε > 0) 
(y : EuclideanSpace ℝ (Fin ↑n))
(ydist : ‖x - y‖ ≤ ε / ↑↑n / μ) 
(h_original_goal : |μ * $\sum$ x, |y x| - μ * $\sum$ x_1, |x x_1|| ≤ ε ) :
|μ * $\sum$ x_1, (|y x_1| - |x x_1|)| ≤ ε := sorry
\end{lstlisting}
\end{tcolorbox}
\begin{tcolorbox}[breakable, enhanced jigsaw, top=-5pt, bottom=-5pt,title=Additional Theorems]
\begin{lstlisting}[style=lean, frame=none, mathescape=true, basicstyle=\small]
theorem norm_one_proximal 
(lasso : h = fun y => μ • ‖y‖₁)
(x : EuclideanSpace ℝ (Fin n)) 
(xm : EuclideanSpace ℝ (Fin n)) 
(tpos : 0 < t) 
(μpos : 0 < μ)
(minpoint : ∀ i : Fin n, xm i = Real.sign (x i) * (max (abs (x i) - t * μ) 0)) :
prox_prop (t • h) x xm
theorem map_map 
(g : S →+* T) 
(f : R →+* S) : 
(s.map f).map g = s.map (g.comp f)
theorem inner_add_left 
(x y z : F) : 
<<x + y, z>> =<<x, z>> + <<y, z>>
theorem mul_pos 
{a b : Ordinal} 
(h₁ : 0 < a) 
(h₂ : 0 < b) : 
0 < a * b
lemma id :
(fst φ).1.comp (inl φ) (add_neg_cancel 1) +
(snd φ).comp (Cochain.ofHom (inr φ)) (add_zero 0) = 
Cochain.ofHom (1 _)
theorem mul_sub 
(a b c : Ordinal) : 
a * (b - c) = a * b - a * c
theorem id : 
(id V).NormNoninc
theorem smul_add 
[Zero R] 
[DistribSMul R V] 
(x : HahnSeries Γ R) 
(y z : HahnModule Γ' R V) :
x • (y + z) = x • y + x • z
theorem add_le_add_iff_right 
{a b : Ordinal} : 
∀ n : ℕ, a + n ≤ b + n ↔ a ≤ b
theorem add_right_cancel 
{a b : Ordinal} 
(n : ℕ) : 
a + n = b + n ↔ a = b
theorem abs_abs 
{x y : R} 
(h : IsCoprime x y) : 
IsCoprime |x| |y|

\end{lstlisting}
\end{tcolorbox}
\begin{tcolorbox}[breakable, enhanced jigsaw, top=-5pt, bottom=-5pt,title=Correct solution without retrieval (STP)]
\begin{lstlisting}[style=lean, frame=none, mathescape=true]
import Mathlib
import Optlib

open Set Real Matrix Finset

open Set Real Matrix Finset NNReal
theorem extracted_formal_statement_162
{n m : ℕ+} 
{A : Matrix (Fin ↑m) (Fin ↑n) ℝ} 
{μ : ℝ}
{μpos : 0 < μ}
{Ane0 : A ≠ 0}
(x : EuclideanSpace ℝ (Fin ↑n))
(a : x ∈ Set.univ)
(ε : ℝ)
(εpos : ε > 0) 
(y : EuclideanSpace ℝ (Fin ↑n)
(ydist : ‖x - y‖ ≤ ε / ↑↑n / μ) 
(h_original_goal : |μ * $\sum$ x, |y x| - μ * $\sum$ x_1, |x x_1|| ≤ ε ) :
|μ * $\sum$ x_1, (|y x_1| - |x x_1|)| ≤ ε := by
  cases' le_total (μ * $\sum$ x_1, |y x_1|) 
    (μ * $\sum$ x_1, |x x_1|) with h h <;>
  simp_all [abs_of_nonneg, abs_of_nonpos, mul_add, mul_sub,      sub_add_eq_add_sub]
  <;> linarith

\end{lstlisting}
\end{tcolorbox}
\begin{tcolorbox}[breakable, enhanced jigsaw, top=-5pt, bottom=-5pt,title=Incorrect solution with retrieval (STP)]
\begin{lstlisting}[style=lean, frame=none, mathescape=true, basicstyle=\small]
import Mathlib
import Optlib

open Set Real Matrix Finset

open Set Real Matrix Finset NNReal

theorem extracted_formal_statement_162 
{n m : ℕ+} 
{A : Matrix (Fin ↑m) (Fin ↑n) ℝ}
{μ : ℝ} 
{μpos : 0 < μ}
{Ane0 : A ≠ 0}
(x : EuclideanSpace ℝ (Fin ↑n)) 
(a : x ∈ Set.univ)
(ε : ℝ)
(εpos : ε > 0) 
(y : EuclideanSpace ℝ (Fin ↑n))
(ydist : ‖x - y‖ ≤ ε / ↑↑n / μ)
(h_original_goal : |μ * $\sum$ x, |y x| - μ * $\sum$ x_1, |x x_1|| ≤ ε) :
|μ * $\sum$ x_1, (|y x_1| - |x x_1|)| ≤ ε := by

  have h₀ : (ε / ↑↑n / μ) * μ ≤ ε := by
    have h₁ : 0 < (↑↑n : ℝ) := by exact_mod_cast 
      (Nat.pos_of_ne_zero (fun h ↦ by simp_all))
    have h₂ : 0 < μ := by exact_mod_cast μpos
    field_simp [h₁, h₂]
    <;> rw [div_le_iff (by positivity)] <;> nlinarith

  have h₁ : |μ * $\sum$ x_1, (|y x_1| - |x x_1|)| ≤ 
    μ * $\sum$ x_1, |y x_1 - x x_1| := by
    have h₂ : ∀ x_1, |μ * (|y x_1| - |x x_1|)| = 
    μ * |y x_1 - x x_1| := by
      intro x_1
      rw [abs_mul, abs_of_pos (by exact_mod_cast μpos)]
    simp_all [abs_sum]

  have h₂ : μ * $\sum$ x_1, |y x_1 - x x_1| ≤ ε := by
    calc
      μ * $\sum$ x_1, |y x_1 - x x_1| ≤ μ * $\sum$ x_1, ‖x - y‖ := by
        exact mul_le_mul_of_nonneg_left (by
          exact Finset.sum_le_sum (fun i _ ↦ abs_le_norm 
            (y i - x i))) (by positivity)
      _ = (μ * $\sum$ x_1 : Fin ↑n, ‖x - y‖) := by 
        simp_all [Finset.sum_const]
      _ = (↑↑n * ‖x - y‖) * μ := by ring
      _ ≤ (↑↑n * (ε / ↑↑n / μ)) * μ := by gcongr
      _ = ε := by field_simp [npos, μpos] <;> ring

  linarith
  <;> simp_all
  <;> linarith

\end{lstlisting}
\end{tcolorbox}
\subsection{Cases of Long-CoT}

We present the output results of the Long-CoT model on our benchmark. The model exhibits redundant responses and suboptimal performance, which stems from issues such as hallucination and overthinking.

\begin{tcolorbox}[breakable, enhanced jigsaw, top=-5pt, bottom=-5pt,title=Problem]
\begin{lstlisting}[style=lean, frame=none, mathescape=true, basicstyle=\small]
theorem extracted_formal_statement_1.{u_3, u_2, u_1} 
{E₁ : Type u_1} 
{E₂ : Type u_2} 
{F : Type u_3}
[inst : NormedAddCommGroup E₁] 
[inst_1 : InnerProductSpace ℝ E₁] 
[inst_2 : FiniteDimensional ℝ E₁]
[inst_3 : NormedAddCommGroup E₂] 
[inst_4 : InnerProductSpace ℝ E₂] 
[inst_5 : FiniteDimensional ℝ E₂]
[inst_6 : NormedAddCommGroup F] 
[inst_7 : InnerProductSpace ℝ F] 
[inst_8 : FiniteDimensional ℝ F]
{admm : ADMM E₁ E₂ F} 
{admm_kkt : Existance_of_kkt admm}
(n : ℕ+) 
⦃xx : E₁⦄ 
(a : xx ∈ univ) ⦃yy : E₁⦄
(a_1 : yy ∈ univ) 
⦃aa bb : ℝ⦄ 
(a_2 : 0 ≤ aa) 
(a_3 : 0 ≤ bb) 
(abh : aa + bb = 1)
(h_original_goal :
aa • (OptProblem.A₁ E₂) xx + bb • (OptProblem.A₁ E₂) yy + 
(OptProblem.A₂ E₁) (ADMM.x₂ E₁ F n.natPred) - 
OptProblem.b E₁ E₂ =
aa • (OptProblem.A₁ E₂) xx + bb • (OptProblem.A₁ E₂) yy +
(aa + bb) • ((OptProblem.A₂ E₁) (ADMM.x₂ E₁ F n.natPred) - 
OptProblem.b E₁ E₂)) :
aa • (OptProblem.A₁ E₂) xx + bb • (OptProblem.A₁ E₂) yy + 
(OptProblem.A₂ E₁) (ADMM.x₂ E₁ F n.natPred) -
OptProblem.b E₁ E₂ =
aa • (OptProblem.A₁ E₂) xx + bb • (OptProblem.A₁ E₂) yy +
1 • ((OptProblem.A₂ E₁) (ADMM.x₂ E₁ F n.natPred) - 
OptProblem.b E₁ E₂) := by

\end{lstlisting}
\end{tcolorbox}

\begin{tcolorbox}[breakable,title=Model Response (Kimina-Prover-Preview-7B)]

\begin{lstlisting}[style=prompt, basicstyle=\small]
# Solving the Formal Statement in Lean 4

I'm going to solve the problem of proving that 
`aa • (OptProblem.A₁ E₂) xx + bb • (OptProblem.A₁ E₂) yy + (OptProblem.A₂ E₁) (ADMM.x₂ E₁ F n.natPred) - 
  OptProblem.b E₁ E₂ = 
aa • (OptProblem.A₁ E₂) xx + bb • (OptProblem.A₁ E₂) yy + 
  1 • ((OptProblem.A₂ E₁) (ADMM.x₂ E₁ F n.natPred) - OptProblem.b E₁ E₂)`

given the assumptions in the problem. 

Let's think about this algebraically first. We have:
- `aa + bb = 1` (from the hypothesis `abh`)
- `0 ≤ aa` and `0 ≤ bb` (from `a_2` and `a_3`)
- The equation to prove is of the form: 
  `x + y + (1 • z) = x + y + z`, where `x = aa • (OptProblem.A₁ E₂) xx`, `y = bb • (OptProblem.A₁ E₂) yy`, and `z = (OptProblem.A₂ E₁) (ADMM.x₂ E₁ F n.natPred) - OptProblem.b E₁ E₂`

This is just algebraic manipulation. Since `1 • z = z` for any scalar `z`, the equation simplifies to showing that 
`x + y + (aa + bb) • z = x + y + z` where `z = (OptProblem.A₂ E₁) (ADMM.x₂ E₁ F n.natpred) - OptProblem.b E₁ E₂`

But since `aa + bb = 1`, we have:
`x + y + (aa + bb) • z = x + y + 1 • z = x + y + z`

So the left side of the equation becomes `x + y + z`, which is exactly the right side. 

In Lean, we can prove this directly from the hypothesis `h_original_goal` by simplifying the expression. Let's complete the proof:

import Mathlib
import Optlib


open Set InnerProductSpace Topology Filter InnerProduct

open scoped Pointwise
theorem extracted_formal_statement_1.{u_3, u_2, u_1} {E₁ : Type u_1} {E₂ : Type u_2} {F : Type u_3}
  [inst : NormedAddCommGroup E₁] [inst_1 : InnerProductSpace ℝ E₁] [inst_2 : FiniteDimensional ℝ E₁]
  [inst_3 : NormedAddCommGroup E₂] [inst_4 : InnerProductSpace ℝ E₂] [inst_5 : FiniteDimensional ℝ E₂]
  [inst_6 : NormedAddCommGroup F] [inst_7 : InnerProductSpace ℝ F] [inst_8 : FiniteDimensional ℝ F]
  {admm : ADMM E₁ E₂ F} {admm_kkt : Existence_of_kkt admm} (n : ℕ+) ⦃xx : E₁⦄ (a : xx ∈ univ) ⦃yy : E₁⦄
  (a_1 : yy ∈ univ) ⦃aa bb : ℝ⦄ (a_1 : 0 ≤ aa) (a_2 : 0 ≤ bb) (abh : aa + bb = 1)
  (h_original_goal :
    aa • (OptProblem.A₁ E₂) xx + bb • (OptProblem.A₁ E₂) yy + (OptProblem.A₂ E₁) (ADMM.x₂ E₁ F n.natPred) -
        OptProblem.b E₁ E₂ =
      aa • (OptProblem.A₁ E₂) xx + bb • (OptProblem.A₁ E₂) yy +
        (aa + bb) • ((OptProblem.A₂ E₁) (ADMM.x₂ E₁ F n.natpred) - OptProblem.b E₁ E₂)) :
  aa • (OptProblem.A₁ E₂) xx + bb • (OptProblem.A₁ E₂) yy + (OptProblem.A₂ E₁) (ADMM.x₂ E₁ F n.natpred) -
      OptProblem.b E₁ E₂ =
    aa • (OptProblem.A₁ E₂) xx + bb • (OptProblem.A₁ E₂) yy +
      1 • ((OptProblem.A₂ E₁) (ADMM.x₂ E₁ F n.natpred) - OptProblem.b E₁ E₂) := by

Now we have the hypothesis `h_original_goal` which states that:
`aa • (A₁ E₂) xx + bb • (A₁ E₂) yy + (A₂ E₁) (x₂ E₁ F n.natpred) - b E₁ E₂ = 
  aa • (A₁ E₂) xx + bb • (A₁ E₂) yy + (aa + bb) • ((A₂ E₁) (x₂ E₁ F n.natpred) - b E₁ E₂)`

We need to prove that:
`aa • (A₁ E₂) xx + bb • (A₁ E₂) yy + (A₂ E₁) (x₂ E₁ F n.natpred) - b E₁ E₂ = 
  aa • (A₁ E₂) xx + bb • (A₁ E₂) yy + 1 • ((A₂ E₁) (x₂ E₁ F n.natpred) - b E₁ E₂)`

From algebra, we know that if `aa + bb = 1`, then `(aa + bb) • z = 1 • z` for any scalar `z`. So we can rewrite the right side of `h_original_goal` as:
`aa • (A₁ E₂) xx + bb • (A₁ E₂) yy + (aa + bb) • ((A₂ E₁) (x₂ E₁ F n.natpred) - b E₁ E₂) = 
  aa • (A₁ E₂) xx + bb • (A₁ E₂) yy + 1 • ((A₂ E₁) (x₂ E₁ F n.natpred) - b E₁ E₂)`

And the left side is exactly what we need to prove. So we can rewrite the goal using the given hypothesis:

have h1 : aa + bb = 1 := abh

Now we can rewrite the goal using the equation in `h_original_goal` and the fact that `aa + bb = 1`:

From `h_original_goal`:
`aa • (A₁ E₂) xx + bb • (A₁ E₂) yy + (A₂ E₁) (x₂ E₁ F n.natpred) - b E₁ E₂ = 
  aa • (A₁ E₂) xx + bb • (A₁ E₂) yy + (aa + bb) • ((A₂ E₁) (x₂ E₁ F n.natpred) - b E₁ E₂)`

Substituting `h1` into the right side:
`aa • (A₁ E₂) xx + bb • (A₁ E₂) yy + (aa + bb) • ((A₂ E₁) (x₂ E₁ F n.natpred) - b E₁ E₂) = 
  aa • (A₁ E₂) xx + bb • (A₁ E₂) yy + 1 • ((A₂ E₁) (x₂ E₁ F n.natpred) - b E₁ E₂)`

So the goal becomes proving:
`aa • (A₁ E₂) xx + bb • (A₁ E₂) yy + (A₂ E₁) (x₂ E₁ F n.natpred) - b E₁ E₂ = 
  aa • (A₁ E₂) xx + bb • (A₁ E₂) yy + 1 • ((A₂ E₁) (x₂ E₁ F n.natpred) - b E₁ E₂)`

Which is exactly what `h_original_goal` states with the substitution of `h1`. Let's complete the proof:

linarith [h_original_goal, h1]

The `linarith` tactic is powerful enough to solve goals involving linear arithmetic. Here, it's using `h_original_goal` which relates the two sides of our goal via the equality given, and `h1` which states that `aa + bb = 1`, to rewrite the goal directly. 

The key insight is that:
- Left side of goal: `aa • (A₁ E₂) xx + bb • (A₁ E₂) yy + (A₂ E₁) (x₂ E₁ F n.natpred) - b E₁ E₂`
- Right side of goal after substitution: `aa • (A₁ E₂) xx + bb • (A₁ E₂) yy + 1 • ((A₂ E₁) (x₂ E₁ F n.natpred) - b E₁ E₂)`

From `h_original_goal`, we have:
`left side = right side` where `right side = left side + (aa + bb) • ((A₂ E₁) (x₂ E₁ F n.natpred) - b E₁ E₂) - 1 • ((A₂ E₁) (x₂ E₁ F n.natpred) - b E₁ E₂)`

But since `aa + bb = 1`, we get:
`left side = left side + 1 • ((A₂ E₁) (x₂ E₁ F n.natpred) - b E₁ E₂) - 1 • ((A₂ E₁) (x₂ E₁ F n.natpred) - b E₁ E₂)`

Which simplifies to:
`left side = left side`

Therefore, `linarith` can directly solve the goal by substituting the given equality and the fact that `aa + bb = 1`.

The complete proof is remarkably concise because the algebraic manipulation is straightforward. The key insight is recognizing that the equation we need to prove is precisely the original goal with the substitution of `aa + bb = 1` into the term involving `(A₂ E₁) (x₂ E₁ F n.natpred) - b E₁ E₂`.

import Mathlib
import Optlib


open Set InnerProductSpace Topology Filter InnerProduct

open scoped Pointwise
theorem extracted_formal_statement_1.{u_3, u_2, u_1} {E₁ : Type u_1} {E₂ : Type u_2} {F : Type u_3}
  [inst : NormedAddCommGroup E₁] [inst_1 : InnerProductSpace ℝ E₁] [inst_2 : FiniteDimensional ℝ E₁]
  [inst_3 : NormedAddCommGroup E₂] [inst_4 : InnerProductSpace ℝ E₂] [inst_5 : FiniteDimensional ℝ E₂]
  [inst_6 : NormedAddCommGroup F] [inst_7 : InnerProductSpace ℝ F] [inst_8 : FiniteDimensional ℝ F]
  {admm : ADMM E₁ E₂ F} {admm_kkt : Existence_of_kkt admm} (n : ℕ+) ⦃xx : E₁⦄ (a : xx ∈ univ) ⦃yy : E₁⦄
  (a_1 : yy ∈ univ) ⦃aa bb : ℝ⦄ (a_1 : 0 ≤ aa) (a_2 : 0 ≤ bb) (abh : aa + bb = 1)
  (h_original_goal :
    aa • (OptProblem.A₁ E₂) xx + bb • (OptProblem.A₁ E₂) yy + (OptProblem.A₂ E₁) (ADMM.x₂ E₁ F n.natPred) -
        OptProblem.b E₁ E₂ =
      aa • (OptProblem.A₁ E₂) xx + bb • (OptProblem.A₁ E₂) yy +
        (aa + bb) • ((OptProblem.A₂ E₁) (ADMM.x₂ E₁ F n.natpred) - OptProblem.b E₁ E₂)) :
  aa • (OptProblem.A₁ E₂) xx + bb • (OptProblem.A₁ E₂) yy + (OptProblem.A₂ E₁) (ADMM.x₂ E₁ F n.natpred) -
      OptProblem.b E₁ E₂ =
    aa • (OptProblem.A₁ E₂) xx + bb • (OptProblem.A₁ E₂) yy +
      1 • ((OptProblem.A₂ E₁) (ADMM.x₂ E₁ F n.natpred) - OptProblem.b E₁ E₂) := by
  have h1 : aa + bb = 1 := abh
  linarith [h_original_goal, h1]
     
\end{lstlisting}
\end{tcolorbox}
\begin{tcolorbox}[breakable,title=Model Response (DeepSeek-Prover-V2 (CoT))]
\begin{lstlisting}[style=prompt, basicstyle=\small]
---

### Detailed Proof and Analysis

#### Understanding the Problem
The given Lean theorem statement is somewhat peculiar because:
1. The hypotheses `a : xx ∈ univ` and `a_1 : yy ∈ univ` are trivial since everything is in the universe (`univ : Set E₁`). Similarly, `abh : aa + bb = 1` is the condition on `aa` and `bb`.
2. The hypothesis `h_original_goal` is exactly the conclusion we need to prove, under certain simplifying assumptions.
3. The goal is to simplify `aa • (OptProblem.A₁ E₂) xx + bb • (OptProblem.A₁ E₂) yy + (aa + bb) • ((OptProblem.A₂ E₁) (ADMM.x₂ E₁ F n.natPred) - OptProblem.b E₁ E₂)` to `aa • (OptProblem.A₁ E₂) xx + bb • (OptProblem.A₁ E₂) yy + 1 • ((OptProblem.A₂ E₁) (ADMM.x₂ E₁ F n.natPred) - OptProblem.b E₁ E₂)` using `abh : aa + bb = 1`.

#### Simplifying the Goal
The expression can be simplified as follows:
`(aa + bb) • ((OptProblem.A₂ E₁) (ADMM.x₂ E₁ F n.natPred) - OptProblem.b E₁ E₂) = 1 • ((OptProblem.A₂ E₁) (ADMM.x₂ E₁ F n.natPred) - OptProblem.b E₁ E₂)` because `aa + bb = 1` by `abh`. 

Thus, the goal is to replace `(aa + bb) • ...` with `1 • ...`, which is true because `(aa + bb) • X = 1 • X` when `aa + bb = 1`.

#### The Proof
1. We know `aa + bb = 1` by `abh`.
2. Thus, `(aa + bb) • ((OptProblem.A₂ E₁) (ADMM.x₂ E₁ F n.natPred) - OptProblem.b E₁ E₂) = 1 • ((OptProblem.A₂ E₁) (ADMM.x₂ E₁ F n.natPred) - OptProblem.b E₁ E₂)` because `aa + bb = 1` and scalar multiplication is linear.
3. Substitute this back into the original equation to get the result.

#### Formal Proof Sketch
1. Replace `(aa + bb) • ((OptProblem.A₂ E₁) (ADMM.x₂ E₁ F n.natPred) - OptProblem.b E₁ E₂)` with `1 • ((OptProblem.A₂ E₁) (ADMM.x₂ E₁ F n.natPred) - OptProblem.b E₁ E₂)` using `abh`.
2. The result is exactly the right-hand side of the goal, so we are done.

#### Abstract Plan
1. Use `abh : aa + bb = 1` to replace `(aa + bb)` with `1` in the scalar multiplication.
2. Simplify the LHS to the RHS:
   - `(aa + bb) • (X) = 1 • (X)`.
3. The rest follows trivially.

---

### Step-by-Step Abstract Plan

1. **Simplify the Scalar Multiplication**:
   - Since `aa + bb = 1`, we have `(aa + bb) • ((OptProblem.A₂ E₁) (ADMM.x₂ E₁ F n.natPred) - OptProblem.b E₁ E₂) = 1 • ((OptProblem.A₂ E₁) (ADMM.x₂ E₁ F n.natPred) - OptProblem.b E₁ E₂)`.
   - This is a direct consequence of the linearity of scalar multiplication and the fact that `aa + bb = 1`.

2. **Substitute into the Hypothesis**:
   - The hypothesis `h_original_goal` is exactly the original equation with the LHS rewritten using the simplified scalar multiplication.

3. **Conclude the Goal**:
   - The RHS of `h_original_goal` is exactly the goal, so we are done.

### Lean 4 `have` Statements

theorem extracted_formal_statement_1.{u_3, u_2, u_1} {E₁ : Type u_1} {E₂ : Type u_2} {F : Type u_3}
  [inst : NormedAddCommGroup E₁] [inst_1 : InnerProductSpace ℝ E₁] [inst_2 : FiniteDimensional ℝ E₁]
  [inst_3 : NormedAddCommGroup E₂] [inst_4 : InnerProductSpace ℝ E₂] [inst_5 : FiniteDimensional ℝ E₂]
  [inst_6 : NormedAddCommGroup F] [inst_7 : InnerProductSpace ℝ F] [inst_8 : FiniteDimensional ℝ F]
  {admm : ADMM E₁ E₂ F} {admm_kkt : Existance_of_kkt admm} (n : ℕ+) ⦃xx : E₁⦄ (a : xx ∈ univ) ⦃yy : E₁⦄
  (a_1 : yy ∈ univ) ⦃aa bb : ℝ⦄ (a_2 : 0 ≤ aa) (a_3 : 0 ≤ bb) (abh : aa + bb = 1)
  (h_original_goal :
    aa • (OptProblem.A₁ E₂) xx + bb • (OptProblem.A₁ E₂) yy + (OptProblem.A₂ E₁) (ADMM.x₂ E₁ F n.natPred) -
        OptProblem.b E₁ E₂ =
      aa • (OptProblem.A₁ E₂) xx + bb • (OptProblem.A₁ E₂) yy +
        (aa + bb) • ((OptProblem.A₂ E₁) (ADMM.x₂ E₁ F n.natPred) - OptProblem.b E₁ E₂)) :
  aa • (OptProblem.A₁ E₂) xx + bb • (OptProblem.A₁ E₂) yy + (OptProblem.A₂ E₁) (ADMM.x₂ E₁ F n.natPred) -
      OptProblem.b E₁ E₂ =
    aa • (OptProblem.A₁ E₂) xx + bb • (OptProblem.A₁ E₂) yy +
      1 • ((OptProblem.A₂ E₁) (ADMM.x₂ E₁ F n.natPred) - OptProblem.b E₁ E₂) := by
  have h_main : (aa + bb) • ((OptProblem.A₂ E₁) (ADMM.x₂ E₁ F n.natPred) - OptProblem.b E₁ E₂) = 1 • ((OptProblem.A₂ E₁) (ADMM.x₂ E₁ F n.natPred) - OptProblem.b E₁ E₂) := by sorry
  have h_final : aa • (OptProblem.A₁ E₂) xx + bb • (OptProblem.A₁ E₂) yy + (OptProblem.A₂ E₁) (ADMM.x₂ E₁ F n.natPred) - OptProblem.b E₁ E₂ = aa • (OptProblem.A₁ E₂) xx + bb • (OptProblem.A₁ E₂) yy + 1 • ((OptProblem.A₂ E₁) (ADMM.x₂ E₁ F n.natPred) - OptProblem.b E₁ E₂) := by sorry
  sorry

### Explanation
- `h_main` captures the scalar multiplication simplification using `aa + bb = 1`.
- `h_final` directly substitutes `h_main` into the original hypothesis `h_original_goal` to obtain the goal.
- The final goal is exactly `h_final`, so we use `sorry` to finish the sketch.

### Complete Lean 4 Proof

theorem extracted_formal_statement_1.{u_3, u_2, u_1} {E₁ : Type u_1} {E₂ : Type u_2} {F : Type u_3}
  [inst : NormedAddCommGroup E₁] [inst_1 : InnerProductSpace ℝ E₁] [inst_2 : FiniteDimensional ℝ E₁]
  [inst_3 : NormedAddCommGroup E₂] [inst_4 : InnerProductSpace ℝ E₂] [inst_5 : FiniteDimensional ℝ E₂]
  [inst_6 : NormedAddCommGroup F] [inst_7 : InnerProductSpace ℝ F] [inst_8 : FiniteDimensional ℝ F]
  [inst_6 : NormedAddCommGroup F] [inst_7 : InnerProductSpace ℝ F] [inst_8 : FiniteDimensional ℝ F]
  {admm : ADMM E₁ E₂ F} {admm_kkt : Existance_of_kkt admm} (n : ℕ+) ⦃xx : E₁⦄ (a : xx ∈ univ) ⦃yy : E₁⦄
  (a_1 : yy ∈ univ) ⦃aa bb : ℝ⦄ (a_2 : 0 ≤ aa) (a_3 : 0 ≤ bb) (abh : aa + bb = 1)
  (h_original_goal :
    aa • (OptProblem.A₁ E₂) xx + bb • (OptProblem.A₁ E₂) yy + (OptProblem.A₂ E₁) (ADMM.x₂ E₁ F n.natPred) -
        OptProblem.b E₁ E₂ =
      aa • (OptProblem.A₁ E₂) xx + bb • (OptProblem.A₁ E₂) yy +
        (aa + bb) • ((OptProblem.A₂ E₁) (ADMM.x₂ E₁ F n.natPred) - OptProblem.b E₁ E₂)) :
  aa • (OptProblem.A₁ E₂) xx + bb • (OptProblem.A₁ E₂) yy + (OptProblem.A₂ E₁) (ADMM.x₂ E₁ F n.natPred) -
      OptProblem.b E₁ E₂ =
    aa • (OptProblem.A₁ E₂) xx + bb • (OptProblem.A₁ E₂) yy +
      1 • ((OptProblem.A₂ E₁) (ADMM.x₂ E₁ F n.natPred) - OptProblem.b E₁ E₂) := by
  have h_main : (aa + bb) • ((OptProblem.A₂ E₁) (ADMM.x₂ E₁ F n.natPred) - OptProblem.b E₁ E₂) = 1 • ((OptProblem.A₂ E₁) (ADMM.x₂ E₁ F n.natPred) - OptProblem.b E₁ E₂) := by
    have h₁ : aa + bb = 1 := abh
    rw [h₁]
    <;> simp [add_smul]
    <;> abel
  
  have h_final : aa • (OptProblem.A₁ E₂) xx + bb • (OptProblem.A₁ E₂) yy + (OptProblem.A₂ E₁) (ADMM.x₂ E₁ F n.natPred) - OptProblem.b E₁ E₂ = aa • (OptProblem.A₁ E₂) xx + bb • (OptProblem.A₁ E₂) yy + 1 • ((OptProblem.A₂ E₁) (ADMM.x₂ E₁ F n.natPred) - OptProblem.b E₁ E₂) := by
    calc
      aa • (OptProblem.A₁ E₂) xx + bb • (OptProblem.A₁ E₂) yy + (OptProblem.A₂ E₁) (ADMM.x₂ E₁ F n.natPred) - OptProblem.b E₁ E₂
        = aa • (OptProblem.A₁ E₂) xx + bb • (OptProblem.A₁ E₂) yy + (aa + bb) • ((OptProblem.A₂ E₁) (ADMM.x₂ E₁ F n.natPred) - OptProblem.b E₁ E₂) := by
          rw [h_original_goal]
          <;> abel
        _ = aa • (OptProblem.A₁ E₂) xx + bb • (OptProblem.A₁ E₂) yy + 1 • ((OptProblem.A₂ E₁) (ADMM.x₂ E₁ F n.natPred) - OptProblem.b E₁ E₂) := by
          rw [h_main]
          <;> simp [add_smul]
          <;> abel
  
  exact h_final
\end{lstlisting}
\end{tcolorbox}

% \section{Limitations and Future Work}

%%%%%%%%%%%%%%%%%%%%%%%%%%%%%%%%%%%%%%%%%%%%%%%%%%%%%%%%%%%%

\end{document}